\RequirePackage{fix-cm}
\documentclass[10pt,letterpaper]{jingdong}
\usepackage[utf8]{inputenc}
\usepackage{enumitem}
\usepackage{array,tabularx,colortbl}
\usepackage{algorithm,algorithmic}
\usepackage{wrapfig}
\usepackage{paracol}
\usepackage{needspace}

\renewcommand{\authorfont}{\fontsize{11}{13}\selectfont}
\crefname{section}{Section}{Sections}
\Crefname{section}{Section}{Sections}

\crefname{figure}{Figure}{Figures}
\Crefname{figure}{Figure}{Figures}

\crefname{table}{Table}{Tables}
\Crefname{table}{Table}{Tables}

\crefname{equation}{Equation}{Equations}
\Crefname{equation}{Equation}{Equations}

\definecolor{cvprblue}{rgb}{0.21,0.49,0.74}
\newtheorem{proposition}{Proposition}

\crefname{proposition}{Proposition}{Propositions}
\Crefname{proposition}{Proposition}{Propositions}
\crefname{algorithm}{Algorithm}{Algorithms}
\Crefname{algorithm}{Algorithm}{Algorithms}

\newcommand{\ind}{\mathbf{1}}
\newcommand{\KL}{D_{\mathrm{KL}}}
\newcommand{\softmax}{\operatorname{softmax}}
\newcommand{\argmax}{\operatorname*{arg\,max}}

\title{PixelOne: From Options to Pixels in a Single Forward for GUI Grounding}
\IfFileExists{latexml.sty}{\usepackage{latexml}}{\newif\iflatexml\latexmlfalse}
\iflatexml
\usepackage{authblk}

\author[1,2]{Kanghui~Tian\thanks{Equal contribution}}
\author[1,2,4]{Shuai~Dong\thanks{Equal contribution}}
\author[2,5]{Yuhan~Zhu\thanks{Equal contribution}}
\author[2]{Congcong~Wang\thanks{Project lead}}
\author[2,4,6]{Ji~Zhao}
\author[3,6]{Tianxiang~Jiang}
\author[2,4,7]{Zhixiong~Zhang}
\author[1,2,4]{Shengyuan~Ding}
\author[3]{Yi~Wang\thanks{Corresponding author}}
\author[2]{Jiaqi~Wang\thanks{Project lead}}
\author[3,4]{Yu~Qiao\thanks{Corresponding author}}

\affil[1]{Fudan University}
\affil[2]{JD.com}
\affil[3]{Shanghai Artificial Intelligence Laboratory}
\affil[4]{Shanghai Innovation Institute}
\affil[5]{Nanjing University}
\affil[6]{University of Science and Technology of China}
\affil[7]{Shanghai Jiao Tong University}
\else

\newcommand{\authorrowseparator}{, }
\newcommand{\authorrowbreak}{\renewcommand{\authorrowseparator}{\protect\\}}
\renewcommand{\author}[2][]{\addtolist[#1]{#2}{\authorlist}{\authorformat}{\authorrowseparator}\renewcommand{\authorrowseparator}{, }}
\newcommand{\affiliationrowseparator}{, }
\newcommand{\affiliationrowbreak}{\renewcommand{\affiliationrowseparator}{\protect\\}}
\renewcommand{\affiliationformat}[2][]{\mbox{{\affiliationfont{\jdregularfont $^{#1}$#2}}}}
\renewcommand{\affiliation}[2][]{\addtolist[#1]{#2}{\affiliationlist}{\affiliationformat}{\affiliationrowseparator}\renewcommand{\affiliationrowseparator}{, }}

\author[1,2,*]{Kanghui~Tian}
\author[1,2,4,*]{Shuai~Dong}
\author[2,5,*]{Yuhan~Zhu}
\author[2,\dagger]{Congcong~Wang}
\authorrowbreak
\author[2,4,6]{Ji~Zhao}
\author[3,6]{Tianxiang~Jiang}
\author[2,4,7]{Zhixiong~Zhang}
\author[1,2,4]{Shengyuan~Ding}
\authorrowbreak
\author[3,\ddagger]{Yi~Wang}
\author[2,\dagger]{Jiaqi~Wang}
\author[3,4,\ddagger]{Yu~Qiao}

\affiliation[1]{Fudan University}
\affiliation[2]{JD.com}
\affiliation[3]{Shanghai Artificial Intelligence Laboratory}
\affiliationrowbreak
\affiliation[4]{Shanghai Innovation Institute}
\affiliation[5]{Nanjing University}
\affiliationrowbreak
\affiliation[6]{University of Science and Technology of China}
\affiliation[7]{Shanghai Jiao Tong University}

\contribution[*]{Equal contribution}
\contribution[\dagger]{Project lead}
\contribution[\ddagger]{Corresponding authors}

\fi

\renewcommand{\checkdatalist}{\centering\small
  \href{https://tiankanghui.github.io/PixelOne/}{%
    \raisebox{-0.15em}{\includegraphics[height=1.05em]{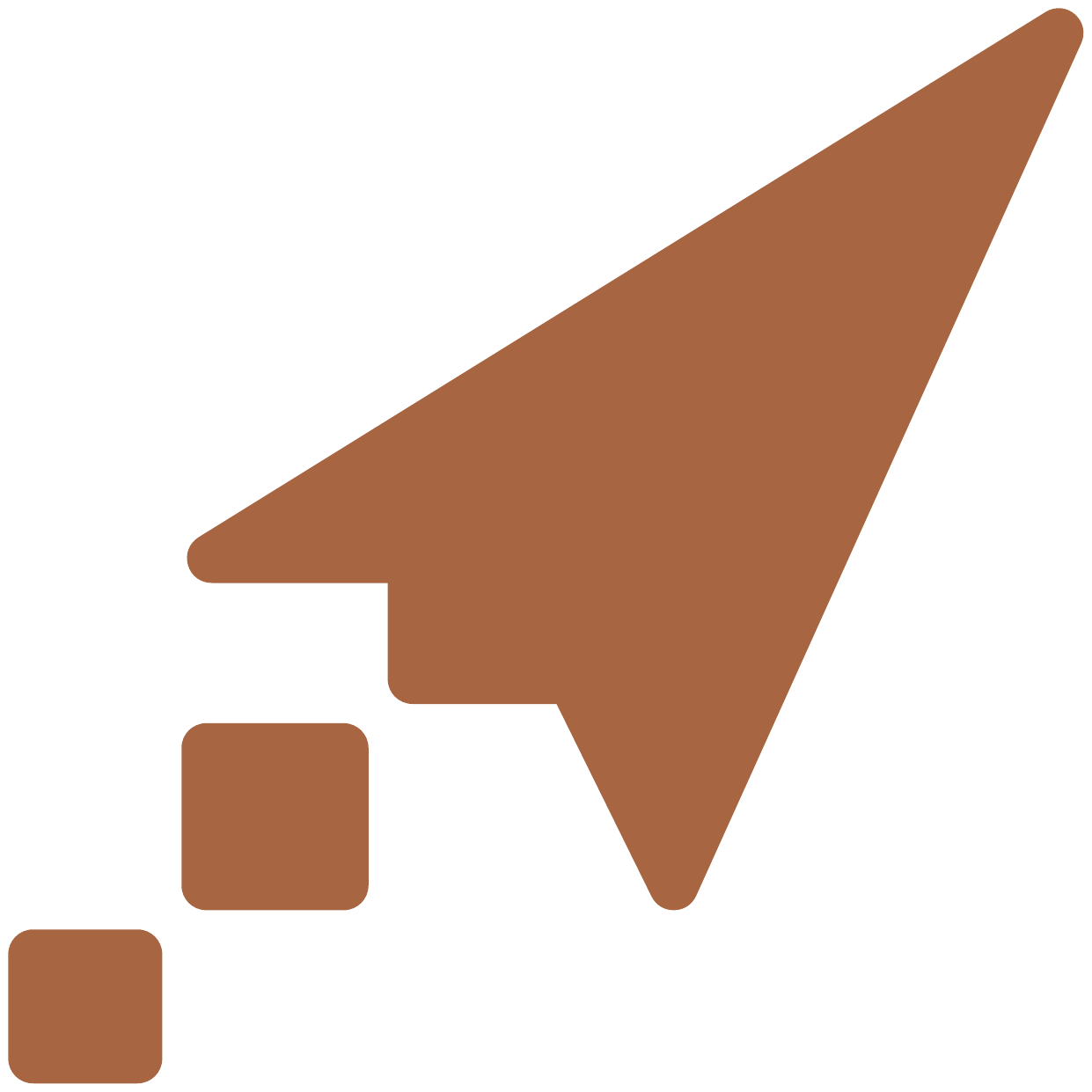}}%
    \hspace{0.35em}\textbf{Page}}
  \qquad
  \href{https://github.com/tiankanghui/PixelOne}{%
    \raisebox{-0.2em}{\includegraphics[height=1.15em]{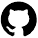}}%
    \hspace{0.4em}\textbf{Code}}
  \qquad
  \mbox{%
    \href{https://huggingface.co/KangHui2200/PixelOne}{%
      \raisebox{-0.2em}{\includegraphics[height=1.15em]{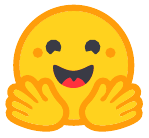}}%
      \hspace{0.4em}\textbf{Models}}%
    \hspace{0.3em}/\hspace{0.3em}%
    \href{https://huggingface.co/datasets/KangHui2200/PixelOne-Data}{%
      \textbf{Data}}%
  }
  \par
}

\abstract{
GUI grounding requires precise clicks at low latency. Recent models such as GUI-Decisions and GUI-Swift predict coordinates in one forward pass using fixed global bins with local refinement. However, screenshot downsampling removes detail needed to locate small controls, while fixed bins cover larger pixel ranges at higher resolutions. We introduce \textbf{PixelOne}, which transfers spatial knowledge from a search teacher to a student that predicts clicks directly from screenshots at native resolution. The teacher uses VideoJEV's text-option scorer to select rows and columns and progressively zoom into candidate regions. The student inherits the teacher's parameters and matches its distributions over explored regions through spatial distillation alongside box supervision. At inference, the same decision head selects a visual token, and a lightweight pixel head predicts the click within it, all in one forward pass. A click-aligned loss encourages pixels inside the target box to score above those outside. Across four benchmarks, PixelOne-Pro-35B-A3B achieves 83.8\% mean accuracy, the highest among the evaluated single-forward grounders, with a mean latency of 0.43\,s per target.
}

\hypersetup{
  pdftitle={PixelOne: From Options to Pixels in a Single Forward for GUI Grounding},
  pdfauthor={Kanghui Tian, Shuai Dong, Yuhan Zhu, Congcong Wang, Ji Zhao, Tianxiang Jiang, Zhixiong Zhang, Shengyuan Ding, Yi Wang, Jiaqi Wang, Yu Qiao},
  pdfsubject={GUI grounding with spatial search distillation and direct pixel prediction}
}

\begin{document}
\maketitle

\begin{figure}[htbp]
\centering
\includegraphics[width=\textwidth]{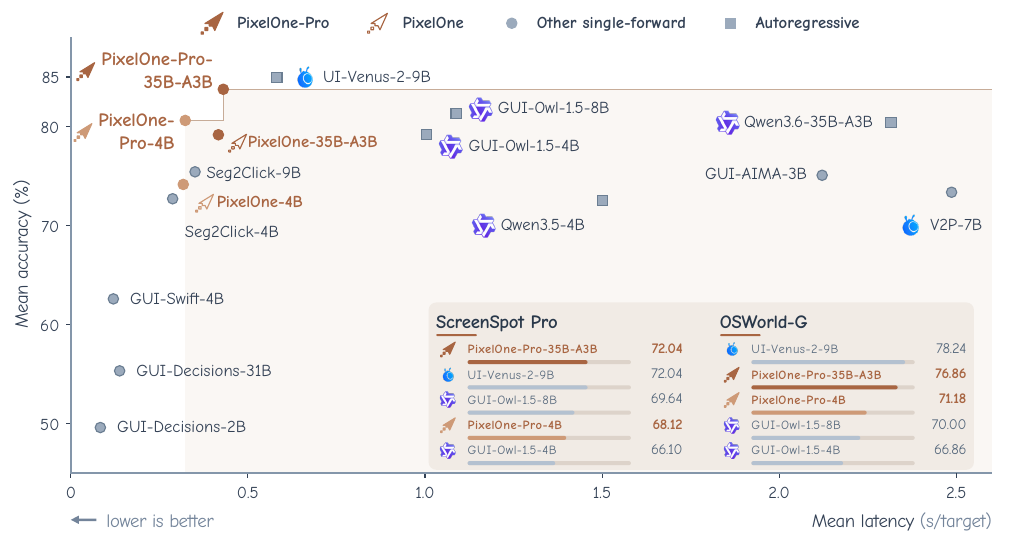}
\caption{\textbf{PixelOne-Pro on the accuracy--latency frontier.} The main plot compares mean accuracy and latency across four benchmarks. The ScreenSpot Pro and OSWorld-G insets each compare both Pro models with the four highest-scoring external GUI-specialized baselines. The leaderboard bars use a truncated linear accuracy scale from 50\% to 80\%. Full results and timing details appear in \cref{tab:main} and \cref{app:timing}.}
\label{fig:teaser}
\end{figure}

\section{Introduction}
\label{sec:intro}

A GUI agent translates instructions, such as ``open the alignment menu'', into corresponding click locations on a screenshot \citep{cua-master, agashe2024agents, xu2025aguvis, he2024webvoyager, xie2024osworld, rawles2025androidworld, xue2026evocua, awadallah2026fara15, yuan2026osworld2, lv2026scalecua, qiu2026vista}. This requires accurate visual grounding, particularly in high-resolution interfaces where small interactive targets are often surrounded by visually similar controls, making the preservation of fine-grained spatial details essential \citep{li2025screenspotpro, hong2024cogagent, zhang2026mvp}. Consequently, a practical GUI agent is required to localize target elements accurately with low latency \citep{aggarwal2026cuaspeedrun}.

Spatial readouts predict clicks directly from scores over image regions or target masks \citep{wu2025guiactor,wang2026seg2click}. By contrast, fixed-slot methods such as GUI-Decisions and GUI-Swift predict distributions over global coordinate bins at predefined token positions and refine the coordinates using neighboring-bin probabilities \citep{umapathi2026slotreading,guiswift2026release}. Both approaches avoid autoregressive coordinate generation and ground targets in a single forward pass. However, these models downsample high-resolution screenshots, removing details needed to locate small controls, while their global bins cover large pixel ranges at higher resolutions. These limitations motivate a direct readout that preserves image detail and predicts clicks at pixel granularity. Native decision models provide a starting point through an interface for scoring candidate options \citep{geuter2026releasing,duan2026general,zhu2026interndecision, cheng2026thisthat, yu2026visualjev}. VideoJEV uses this interface to select textual answers and temporal intervals \citep{zhu2026videojev}. Extending it to pixel localization requires spatial candidates tied to screenshot regions and a way to resolve a click within the selected region.

\begin{figure*}[t]
\centering
\includegraphics[width=\textwidth]{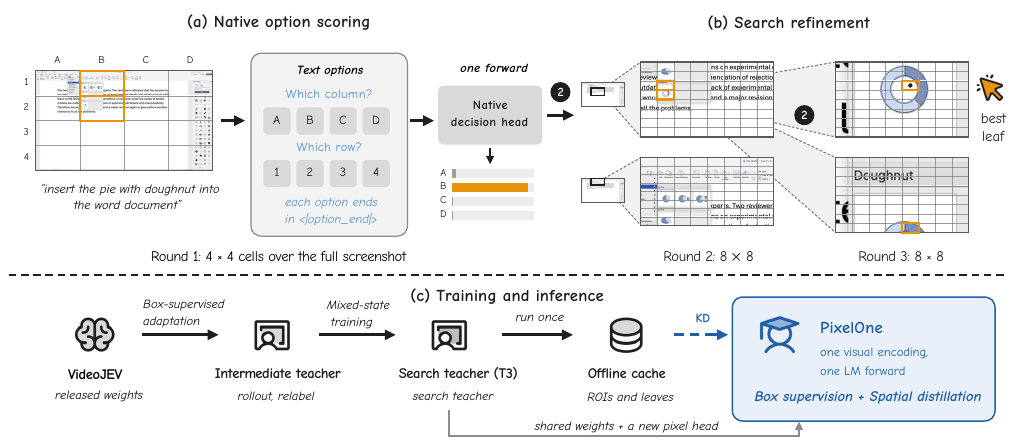}
\caption{\textbf{From spatial search to single-forward grounding.} (a) The teacher scores row and column choices with VideoJEV's native head. (b) The teacher refines candidate regions through zoomed search. (c) The student inherits the teacher's shared parameters and learns from its spatial distributions to predict a click with one visual encoding and one model forward.}
\label{fig:overview}
\end{figure*}

We introduce \textbf{PixelOne}, a teacher-student framework for accurate GUI grounding from full-resolution screenshots in a single forward pass. During training, a search teacher learns to localize targets by choosing rows and columns from textual spatial options. It progressively narrows the target location across full-image and zoomed views using box supervision (\cref{fig:overview}).
The student inherits the teacher's network and query/key decision head. The teacher also provides distributions over the regions explored during search, which complement the student's box supervision. This transfer is possible because the same head scores both forms of spatial evidence (\cref{fig:readout}): textual options in the teacher and visual tokens in the student.
At inference, the student applies the shared decision head to visual tokens from the full-resolution screenshot and selects the token most relevant to the instruction.

PixelOne determines the precise click within that region as the selected token localizes the target only to an area. A pixel head predicts this point using features from the same forward pass. Assigning high probability to pixels inside the target does not guarantee the top-ranked pixel is a valid click. We therefore introduce a click-aligned objective that encourages valid pixels to score higher than pixels outside the target. PixelOne returns a click location with one visual encoding and one pass through the language model, without iterative search or autoregressive coordinate generation.

We train PixelOne in two settings: the standard version updates language LoRA and both readout heads, whereas PixelOne-Pro updates the entire model. Across four benchmarks, PixelOne-Pro-35B-A3B achieves 83.8\% mean accuracy at an average latency of 0.43\,s per click, ranking first among the evaluated single-forward grounders. Pro-4B reaches 80.6\% at 0.32\,s, and both models lie on the accuracy--latency frontier (\cref{tab:main,fig:teaser}). Ablations across both model scales demonstrate the benefits of spatial distillation and the click-aligned objective (\cref{tab:ablation,tab:winner,tab:fraction}).

Our contributions are:
\begin{itemize}
\item \textbf{Native pixel-decision interface.} We adapt VideoJEV's head to visual tokens and resolve clicks with a lightweight pixel head trained with a click-aligned objective.
\item \textbf{Search-to-pixel learning.} We share decision parameters across textual search options and visual tokens, transferring both the teacher's weights and its spatial distributions to the student.
\item \textbf{Accurate and efficient grounding.} PixelOne grounds full-resolution screenshots in one forward pass at 4B and 35B-A3B, with both Pro variants on the evaluated accuracy--latency frontier.
\end{itemize}

\section{Related Work}
\label{sec:related}
\paragraph{Coordinate generation and direct readout.}
Autoregressive grounders represent coordinates and action fields as text \citep{cheng2024seeclick,qin2025uitars,xu2026mobileagent35,venusteam2026uivenus2technicalreport}. This flexible interface requires sequential decoding, whereas direct readouts predict locations in a single forward pass. GUI-Actor and V2P score visual patches with a dedicated head, while GUI-AIMA supervises attention within the backbone \citep{wu2025guiactor,chen2025v2p,zhou2025guiaima}. Seg2Click instead predicts action masks at the pixel level \citep{wang2026seg2click}. Fixed-slot methods such as GUI-Decisions and GUI-Swift take another route, reading coordinates from predefined positions through the vocabulary head \citep{umapathi2026slotreading,guiswift2026release}. Although these methods avoid sequential decoding, precise grounding still depends on sufficient image detail and reliably converting spatial scores into clicks.

\paragraph{Spatial detail and inference cost.}
Higher-resolution screenshots preserve the detail needed to locate small targets but increase inference cost \citep{fan2026rankground}. GUI-Decisions and GUI-Swift limit the number of visual tokens and therefore downsample large screenshots \citep{umapathi2026slotreading,guiswift2026release}. They also predict coordinates using fixed global bins with refinement from neighboring-bin probabilities; as resolution increases, each bin covers a larger pixel range. Other methods recover local detail through additional computation. Zoomed search processes cropped regions, while Re-Prefill uses attention to guide a second prefill stage \citep{li2025screenspotpro,lin2026prefill}. InnerZoom avoids another round of image encoding by refining regional evidence within the decoder layers, although it still generates coordinates autoregressively \citep{liu2026innerzoom}. An alternative is to move spatial refinement into training, allowing the model to learn from detailed local views without repeating the process at inference time.

\paragraph{Transferring search knowledge to direct prediction.}
Methods that transfer knowledge from refined views differ in what they teach the student. Region-to-Image Distillation turns zoomed regions into question and answer pairs for models that process the full image \citep{wei2026zooming}. Native decision models provide a different signal by assigning probabilities to a set of options \citep{geuter2026releasing,duan2026general,zhu2026interndecision,qin2026mjev}. These options may represent actions from an accessibility tree, as in Jev-Mobile, or regions proposed by a segmenter, as in AffordJEV \citep{zhang2026jevmobile,li2026affordjev}. VideoJEV further shows that the same scoring interface can evaluate textual answers and temporal intervals from video evidence \citep{zhu2026videojev}. PixelOne uses this interface for spatial search and transfers both the learned parameters and the resulting distributions over image regions to a direct predictor. The student then scores visual tokens and predicts a precise click from the full-resolution screenshot in one forward pass.

\section{Method}
\label{sec:method}
PixelOne takes a screenshot $I$ and an instruction $u$ as input and predicts a click location $\hat p$. The prediction is considered correct if $\hat p$ lies inside the target box $B$. Our goal is to predict this point directly from the full screenshot while learning from the finer spatial decisions made during search.

\begin{figure*}[t]
\centering
\includegraphics[width=\textwidth]{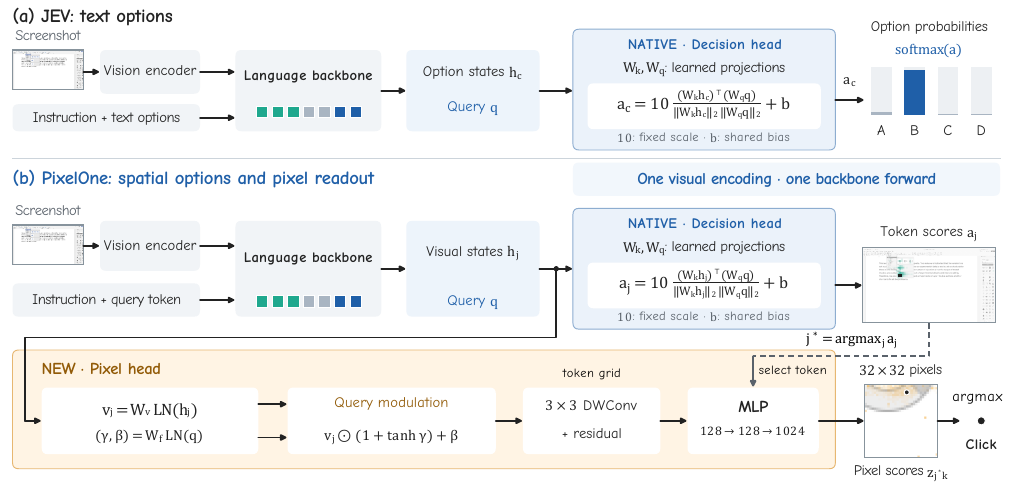}
\caption{\textbf{One decision head, two kinds of options.} (a) The teacher uses the native decision head to score text options. (b) PixelOne reuses this head to select a visual token and adds a pixel head to locate the click within it.}
\label{fig:readout}
\end{figure*}

To this end, we first train a teacher to locate the target through a sequence of row and column choices (\cref{fig:overview}). We then transfer its knowledge to a student that operates without search. The student uses the teacher's decision head to select a visual token and a lightweight pixel head to determine the click within that token (\cref{fig:readout}). It inherits the teacher's parameters and learns from both the teacher's spatial distributions and the target box. At inference time, the student receives only $(I,u)$ and predicts the click with one visual encoding and one pass through the model.

\subsection{From Native Options to Spatial Localization}
\label{sec:interface}
\label{sec:teacher}
VideoJEV's native interface scores candidate options rather than predicting image coordinates directly \citep{zhu2026videojev}. We adapt this interface to spatial localization by representing a target through row and column choices on a grid. Applying the same choices to progressively zoomed crops gradually narrows the target region.

A single forward pass scores all questions associated with one search state. For each question, let $q$ denote the hidden state after the option list and $h_c$ the hidden state at the end of candidate $c$. The native query/key head assigns candidate $c$ the score
\begin{equation}
a_c=\alpha\frac{(W_kh_c)^\top(W_qq)}
{\|W_kh_c\|\,\|W_qq\|}+b.
\label{eq:coarse}
\end{equation}
Here $W_q q$ and $W_k h_c$ denote the native affine projections, with their biases implicit. The scale $\alpha=10$ and bias $b$ remain fixed during teacher training. Softmax is applied separately to the row, column, and visibility option sets.

The three-round search (T3) begins with a $4\times4$ grid over the full screenshot. The score of each cell is the product of its row and column probabilities. The teacher retains the two cells with the highest scores.

In each of the next two rounds, the teacher places an $8\times8$ grid over a crop corresponding to each retained cell. For every crop, it receives both the enlarged crop and a full-image overview that marks the crop location. It also answers a visibility question that asks whether the target is present in the crop. A child score is obtained by multiplying its row and column probabilities by the visibility probability of the crop and the score of its parent. The teacher then retains the two highest-scoring children across both crops. After the third round, it returns the center of the best final cell as $\hat p_T$. The two crop states in each refinement round are processed as a batch, resulting in three language-model calls and five visual-encoder calls per target (\cref{fig:overview}).

We first train the teacher on global and local states derived from target boxes. We then run the partially trained teacher on a subset of the training examples and collect the states visited during search. For each visited crop, the center of the target box defines the row and column labels whenever it falls inside the crop. The visibility label is positive when the crop fully contains the target box and negative when the crop does not overlap it. For partial overlaps, we omit the visibility loss while retaining any valid row and column supervision. Training then continues on a mixture of visited states and states derived from target boxes \citep{ross2011reduction}. Finally, we freeze the teacher and cache its search traces (\cref{app:teacher}). The teacher's weights initialize the student, while the cached distributions provide spatial supervision.

\subsection{From Search to Single-Forward Pixels}
\label{sec:student}
The student transfers the learned decision parameters from textual options to visual tokens in the screenshot. Each of the $N$ tokens represents an $s\times s$ image patch, where $s=32$.

\paragraph{Selecting a region.}
A single forward pass produces a state $h_j$ for each visual token and a query state $q$ after the instruction. We apply the decision head in \cref{eq:coarse} to obtain a score $a_j$ for every token. Scoring each token against the query makes the resulting scores depend on the instruction.

\paragraph{Selecting a pixel.}
A visual token identifies only an image region, so a lightweight pixel head predicts the precise point within it. The head uses the same token states, conditions them on the query $q$, and produces a logit $z_{jk}$ for each of the $s^2$ pixel centers $k$ in token $j$ (\cref{app:readout}). Let $V_j$ denote the set of valid pixel indices in token $j$, excluding padding outside the original image. Decoding first selects the token with the highest score and then the highest scoring valid pixel within that token:
\begin{equation}
 j^*=\argmax_j a_j,\qquad
 k^*=\argmax_{k\in V_{j^*}}z_{j^*k}.
 \label{eq:decode}
\end{equation}
The predicted click is $\hat p=p_{j^*k^*}$, where $p_{jk}$ denotes the coordinate of pixel center $k$ in the original image. The native head selects the target region, and the pixel head determines the precise location within it (\cref{fig:readout}). Both operate on features from the same forward pass, without encoding additional crops or generating coordinates autoregressively.

\subsection{Transferring Spatial Knowledge}
\label{sec:kd}

The teacher transfers spatial knowledge to the student in two ways. First, the student inherits the teacher's backbone and native decision head, while its pixel head is initialized separately. Second, the teacher's search traces provide soft spatial targets that describe how probability is distributed over the explored regions.

\paragraph{Turning search traces into spatial targets.}
Each search trace ends in a set of leaf regions $R_\ell$. The corresponding log score $s_\ell$ accumulates the visibility, row, and column scores along the path to leaf $\ell$. We convert these scored regions into a distribution over pixels by spreading each leaf's weight uniformly across its area. When multiple leaves cover the same pixel, their contributions are added:
\begin{equation}
 m_{jk}\propto\sum_\ell e^{s_\ell-\max_r s_r}
           \frac{|P_{jk}\cap R_\ell|}{|R_\ell|},
 \label{eq:project}
\end{equation}
where $P_{jk}$ denotes the unit square associated with pixel $k$ in token $j$. We normalize the resulting values over all pixels in the original image so that $\sum_{j,k}m_{jk}=1$. The teacher mass assigned to token $j$ is then $m_j=\sum_k m_{jk}$. This projection allows the same search trace to supervise both token selection and pixel prediction.

\paragraph{Distilling over observed regions.}
A search trace provides teacher evidence only within the regions covered by its leaves. Pixels outside these regions remain unobserved and should not be treated as negatives for distillation (\cref{fig:supervision}a). We therefore restrict matching to the observed tokens
$O=\{j:m_j>0\}$ and, within each such token, the observed pixels
$O_j=\{k:m_{jk}>0\}$.

Let $m_O=(m_j)_{j\in O}$ denote the teacher masses over observed tokens and
$m_{j,O_j}=(m_{jk})_{k\in O_j}$ the corresponding masses over observed pixels in token $j$.
We soften and normalize a mass vector using
\begin{equation}
 [\mathcal T_T(v)]_i
 =\frac{v_i^{1/T}}{\sum_r v_r^{1/T}}.
 \label{eq:mass_temperature}
\end{equation}
We then define
$\tilde m_O=\mathcal T_T(m_O)$ and
$\tilde m_{j,O_j}=\mathcal T_T(m_{j,O_j})$.
Here the sum runs over the entries of $v$, and we use $T=2$.
The spatial distillation objective is
\begin{equation}
\begin{split}
 \mathcal L_{\mathrm{KD}}=T^2\Big[&
 \KL\big(\tilde m_O\,\|\,\sigma(a_O/T)\big)\\
 +&\sum_{j\in O}m_j\,
 \KL\big(\tilde m_{j,O_j}\,\|\,\sigma(z_{j,O_j}/T)\big)\Big],
\end{split}
 \label{eq:kd}
\end{equation}
where $\sigma$ denotes softmax,
$a_O=(a_j)_{j\in O}$, and
$z_{j,O_j}=(z_{jk})_{k\in O_j}$.
The first term matches the teacher distribution over observed tokens. The second matches the conditional pixel distribution within each observed token and weights it by the token's original mass $m_j$. Region KD provides an alternative that retains the token term but replaces pixel-level matching with leaf-region mass matching (\cref{app:region}).

We apply $\mathcal L_{\mathrm{KD}}$ only when the teacher's final click
$\hat p_T$ lies inside the target box $B$. This gate does not affect box supervision, which is applied to every training example.

\subsection{Learning to Select Valid Clicks}
\label{sec:loss}

A target box contains many valid click locations, whereas inference must return a single point. We therefore supervise token selection and within-token pixel selection separately, using the box to determine the valid choices at each level.

\paragraph{Token-level supervision.}
Let $V_j$ denote the indices of valid image pixels in token $j$, and let $p_{jk}$ be the center of pixel $k$ in original-image. We define
$A_j=\{k\in V_j:p_{jk}\in B\}$ as the target pixels in token $j$ and
$J_B=\{j:A_j\neq\emptyset\}$ as the tokens containing at least one such pixel. The token loss maximizes the probability assigned to these tokens:
\begin{equation}
 \mathcal L_{\mathrm{coarse}}
 =-\log\sum_{j\in J_B}\softmax(a)_j.
 \label{eq:coarseloss}
\end{equation}

\paragraph{Pixel-level winner loss.}
At the pixel level, high total probability on the target pixels does not necessarily produce a valid argmax. For example, ten target pixels with probability $0.09$ each account for $90\%$ of the probability mass, but a non-target pixel with probability $0.10$ still determines the click. We therefore compare each target pixel with all non-target pixels in the same token:
\begin{equation}
 \ell_{jk}=\log\Big(1+\sum_{v\in V_j\setminus A_j}
                  e^{z_{jv}-z_{jk}}\Big).
 \label{eq:winner}
\end{equation}
This loss decreases as target pixel $k$ outranks its non-target competitors. For the selected token $j^*$, if $\ell_{j^*k}<\log 2$ for some $k\in A_{j^*}$, then $\sum_{v\in V_{j^*}\setminus A_{j^*}}e^{z_{j^*v}-z_{j^*k}}<1$. Each term is therefore below one, so every non-target pixel scores below this target pixel and the argmax click lies inside $B$. The loss and decoder use the same valid-image-pixel mask, so this sufficient condition applies directly at inference (\cref{prop:winner,app:proof}).

To learn from multiple valid clicks without fixing a single target point, we apply this loss to the highest-scoring fraction $\rho$ of the target pixels:
\begin{equation}
 \mathcal L_{\mathrm{pix}}(j,B)
 =\frac{1}{|S_j|}\sum_{k\in S_j}\ell_{jk},
 \qquad
 |S_j|=\lceil\rho\,|A_j|\rceil,
 \label{eq:topfrac}
\end{equation}
where $S_j$ contains the $|S_j|$ target pixels with the highest current logits $z_{jk}$. Membership in $S_j$ is selected without differentiation. Gradients flow through the selected target logits and their non-target competitors in \cref{eq:winner}. The main-table PixelOne models use $\rho=0.2$, denoted Top20. Setting $|S_j|=1$ gives the single-winner variant used in \cref{tab:hardkd}, while \cref{app:fraction} also compares these choices with $\rho=1$. A token with no non-target image pixels has zero pixel loss.

\paragraph{Joint objective.}
Let $R\subseteq J_B$ contain the tokens whose pixel logits are read for box supervision (\cref{app:readout}). Additional readouts used for distillation do not change $R$ or the averaging below. We combine the corresponding pixel losses with token-level supervision:
\begin{equation}
 \mathcal L_{\mathrm{GT}}
 =\mathcal L_{\mathrm{coarse}}
 +\frac{1}{|R|}\sum_{j\in R}\mathcal L_{\mathrm{pix}}(j,B).
 \label{eq:gt}
\end{equation}
The complete objective adds spatial distillation when the teacher's final click lies inside the target box:
\begin{equation}
 \mathcal L
 =\mathcal L_{\mathrm{GT}}
 +\lambda\,\ind[\hat p_T\in B]\,\mathcal L_{\mathrm{KD}},
 \label{eq:total}
\end{equation}
where $\lambda=0.5$. Every example contributes box supervision, while a correct teacher click enables the additional spatial distillation term. Omitting this term gives the GT-only objective. \Cref{alg:student} summarizes training and inference.

% \paragraph{Training configurations.}
% All main-table PixelOne variants use the same readout and objective. The controlled configuration trains the language LoRA parameters and both prediction heads while keeping the visual encoder frozen and using causal attention. The initial student inherits the search teacher's trained parameters, while its pixel head is newly initialized. At each scale, PixelOne and PixelOne-Pro continue the same controlled student in separate branches, retaining its trained pixel head. PixelOne-Pro updates all parameters, including the visual encoder, and uses bidirectional masks in the existing full-attention layers. The linear-attention layers retain their original computation (\cref{app:readout,app:pro}).
\paragraph{Model variants.}
PixelOne and the fully fine-tuned PixelOne-Pro share the same readout and training objective. Their optimization details are provided in \cref{app:pro}.

% \begin{algorithm}[!h]
% \caption{PixelOne training and inference}
% \label{alg:student}
% \begin{algorithmic}[1]
% \STATE \textbf{Offline:} Train the T3 search teacher. Cache its traces if KD is used.
% \IF{constructing the initial student}
% \STATE Inherit the teacher's shared parameters and initialize a new pixel head.
% \ELSE
% \STATE Load the corresponding trained student, including its trained pixel head.
% \ENDIF
% \FOR{each training step}
% \STATE Sample $(I,u,B)$.
% \STATE Retrieve its cached teacher trace if training with KD.
% \STATE Run a single forward pass over the screenshot and instruction.
% \STATE Score all visual tokens using \cref{eq:coarse}.
% \STATE Read pixel logits for box supervision and prediction, adding teacher-observed tokens only if training with KD and $\hat p_T\in B$.
% \STATE Set $\mathcal L\leftarrow\mathcal L_{\mathrm{GT}}$.
% \IF{training with KD and $\hat p_T\in B$}
% \STATE Add $\lambda\mathcal L_{\mathrm{KD}}$ to $\mathcal L$.
% \ENDIF
% \STATE Update the trainable parameters of the current configuration.
% \ENDFOR
% \STATE \textbf{Inference:} Run a single forward pass, then select a token and a pixel within it using \cref{eq:decode}.
% \end{algorithmic}
% \end{algorithm}
\begin{algorithm}[!ht]
\caption{PixelOne training and inference}
\label{alg:student}
\begin{algorithmic}[1]
\STATE Train T3; cache traces for KD.
\STATE Initialize from T3, or resume a trained student.
\FOR{each training step}
\STATE Sample $(I,u,B)$; load its KD trace if needed.
\STATE Forward once; score tokens and read GT/pred pixels.
\STATE $\mathcal L\leftarrow\mathcal L_{\mathrm{GT}}$.
\IF{using KD and $\hat p_T\in B$}
\STATE Read trace-observed pixels; add $\lambda\mathcal L_{\mathrm{KD}}$.
\ENDIF
\STATE Update trainable parameters.
\ENDFOR
\STATE \textbf{Inference:} In one pass, select a token and then a pixel.
\end{algorithmic}
\end{algorithm}

\section{Experiments}
\label{sec:experiments}
% \vspace{-0.5em}
We compare PixelOne against existing GUI grounders and Qwen-based baselines \citep{qwen35,qwen36}. We then conduct controlled studies to evaluate search knowledge transfer and the click-aligned pixel objective.

\begin{table*}[t]
\captionsetup{width=\textwidth}
\caption{\textbf{Grounding accuracy and latency compared with external methods.} Accuracy (\%) is evaluated locally on each full benchmark. Mean accuracy and latency weight the four benchmarks equally. Bold and underlined values indicate the best and second-best results in each parameter group. $P$ and $D$ denote prefill and autoregressive decoding calls. Training data are reported in their original adaptation-stage units, and -- indicates an unverified count. $^\dagger$ marks the Pro configuration (\cref{app:pro}). See \cref{app:recentbaselines,app:timing} for model and timing details.}

\label{tab:main}
\centering\footnotesize
\renewcommand{\textbf}[1]{{\bfseries #1}}
\setlength{\tabcolsep}{1.5pt}
\renewcommand{\tabularxcolumn}[1]{m{#1}}
\newcommand{\blk}[1]{\multicolumn{10}{@{}l}{\textit{#1}}\\[1pt]}
\begin{tabularx}{\textwidth}{@{}llcl*{6}{>{\centering\arraybackslash}X}@{}}
\toprule
Method & Params & Calls & Training data & ScreenSpot & \shortstack{ScreenSpot\\V2} & \shortstack{ScreenSpot\\Pro} & OSWorld-G & Avg. & s/target\\
\midrule
\blk{2B--4B models}
GUI-Decisions-2B \citep{umapathi2026slotreading} & 2B & $1P$ & 7K ex. & 76.97 & 79.17 & 13.98 & 28.24 & 49.59 & \textbf{0.083}\\
GUI-AIMA-3B \citep{zhou2025guiaima} & 3B & $1P$ & 509K samp. & 88.68 & 91.67 & 53.89 & 66.08 & 75.08 & 2.121\\
Seg2Click-4B \citep{wang2026seg2click} & 4B & $1P$ & 1.26M rec. & \textbf{91.59} & \textbf{93.32} & 50.22 & 55.69 & 72.70 & 0.287\\
GUI-Owl-1.5-4B \citep{xu2026mobileagent35} & 4B & $1P{+}D$ & -- & \underline{90.72} & \underline{93.16} & \underline{66.10} & \underline{66.86} & \underline{79.21} & 1.004\\
GUI-Swift-4B \citep{guiswift2026release} & 4B & $1P$ & 27.5K ex. & 86.24 & 88.99 & 29.22 & 45.88 & 62.59 & \underline{0.120}\\
Qwen3.5-4B & 4B & $1P{+}D$ & none & 87.81 & 90.64 & 55.03 & 56.67 & 72.54 & 1.501\\
\cmidrule(lr){1-10}
\rowcolor{cvprblue!8}
\textbf{PixelOne} & 4B & $1P$ & 200K tgt. & 90.64 & \underline{93.16} & 55.34 & 57.45 & 74.15 & 0.317\\
\rowcolor{cvprblue!8}
\textbf{PixelOne-Pro}$^\dagger$ & 4B & $1P$ & 500K tgt. & 90.33 & 92.85 & \textbf{68.12} & \textbf{71.18} & \textbf{80.62} & 0.323\\
\midrule
\blk{7B--35B models; 35B-A3B has about 3B active parameters}
V2P-7B \citep{chen2025v2p} & 7B & $1P$ & 9.6M elem. & 90.49 & 92.61 & 52.31 & 58.04 & 73.36 & 2.487\\
GUI-Owl-1.5-8B \citep{xu2026mobileagent35} & 8B & $1P{+}D$ & -- & 91.67 & 94.03 & \underline{69.64} & 70.00 & 81.33 & 1.088\\
Seg2Click-9B \citep{wang2026seg2click} & 9B & $1P$ & 1.26M rec. & \underline{92.14} & 94.42 & 55.34 & 59.80 & 75.43 & \underline{0.351}\\
UI-Venus-2-9B \citep{venusteam2026uivenus2technicalreport} & 9B & $1P{+}D$ & -- & \textbf{93.32} & \textbf{96.23} & \textbf{72.04} & \textbf{78.24} & \textbf{84.96} & 0.582\\
GUI-Decisions-31B \citep{umapathi2026slotreading} & 31B & $1P$ & 21K ex. & 86.64 & 88.68 & 12.02 & 33.92 & 55.31 & \textbf{0.138}\\
Qwen3.6-35B-A3B & 35B-A3B & $1P{+}D$ & none & 91.51 & 93.71 & 59.96 & 76.47 & 80.41 & 2.316\\
\cmidrule(lr){1-10}
\rowcolor{cvprblue!8}
\textbf{PixelOne} & 35B-A3B & $1P$ & 200K tgt. & 91.04 & 93.32 & 63.50 & 68.82 & 79.17 & 0.417\\
\rowcolor{cvprblue!8}
\textbf{PixelOne-Pro}$^\dagger$ & 35B-A3B & $1P$ & 500K tgt. & 91.67 & \underline{94.50} & \textbf{72.04} & \underline{76.86} & \underline{83.77} & 0.431\\
\bottomrule
\end{tabularx}
\end{table*}

\paragraph{Evaluation.}
We report point-in-box accuracy on the full target sets of ScreenSpot \citep{cheng2024seeclick}, ScreenSpot V2 \citep{wu2024osatlas}, and ScreenSpot Pro \citep{li2025screenspotpro}, as well as on the 510 targets with rectangular boxes in OSWorld-G \citep{xie2025jedi}. We average accuracy and latency equally across the four benchmarks and compute speedups from the resulting mean latencies. All models in \cref{tab:main} are evaluated locally using their respective image-processing policies. Latency is measured from receiving a decoded screenshot to producing a click. Further evaluation details are provided in \cref{app:eval}.

\paragraph{Training.}
The controlled studies use 21,557 training targets. At each model scale, PixelOne and PixelOne-Pro branch separately from the corresponding Top20+KD student and are further trained on 200K and 500K sampled targets, respectively. Further details on the data and training configurations are provided in \cref{app:data,app:init}.

\subsection{Main Grounding Results}
\label{sec:external}
\label{sec:mainresults}
\paragraph{Leading single-forward accuracy.}
Across the four benchmarks in \cref{tab:main}, PixelOne-Pro reaches 83.77\% mean accuracy at 35B-A3B and 80.62\% at 4B. Pro-35B-A3B leads the single-forward grounders, outperforming the strongest external baseline, Seg2Click-9B, by 8.34 percentage points.

\paragraph{Faster and more accurate than GUI-Owl-1.5.}
Pro-4B outperforms GUI-Owl-1.5-4B \citep{xu2026mobileagent35} by 1.41 percentage points with a $3.11\times$ speedup. Pro-35B-A3B similarly outperforms GUI-Owl-1.5-8B by 2.43 points with a $2.53\times$ speedup.

\paragraph{A lower-latency alternative to UI-Venus-2.}
Pro-35B-A3B achieves 26\% lower mean latency than UI-Venus-2-9B \citep{venusteam2026uivenus2technicalreport}, with 83.77\% mean accuracy versus 84.96\%. Both reach 72.04\% on ScreenSpot Pro. \Cref{fig:teaser} summarizes these comparisons: the insets show the benchmark-level gains over GUI-Owl-1.5, and the main plot shows both PixelOne-Pro models on the accuracy--latency frontier.

\paragraph{Parameter-efficient grounding.}
With language LoRA and a frozen visual encoder, the 200K PixelOne models reach 74.15\% mean accuracy at 4B and 79.17\% at 35B-A3B.

\begin{figure*}[t]
\centering
\includegraphics[width=\textwidth]{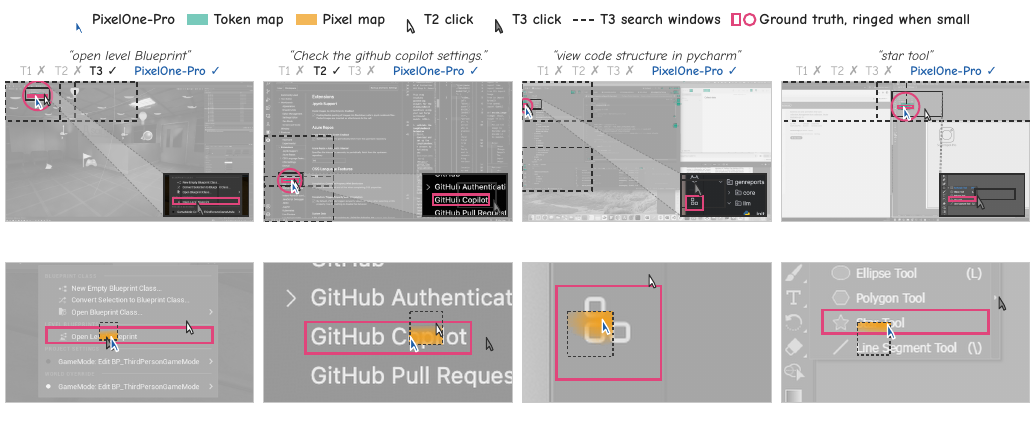}
\caption{\textbf{Direct pixel prediction alongside teacher search.} PixelOne-Pro-35B-A3B locates the shown ScreenSpot Pro targets in a single forward pass. Across the full benchmark, it correctly localizes 230 targets missed by T3. \Cref{app:qual} provides further analysis.}
\label{fig:qualitative}
\end{figure*}

\paragraph{Qualitative results.}
\Cref{fig:qualitative} compares full-screenshot predictions from Pro-35B-A3B with teacher search trajectories.

\subsection{Inference Efficiency}
\label{sec:efficiency}
\label{sec:interfaces}
\paragraph{Eliminating sequential decoding.}
PixelOne produces a click with one visual encoding and one language-model forward, avoiding sequential coordinate decoding. Although UI-Venus-2-9B and both GUI-Owl-1.5 models also encode the screenshot once, UI-Venus-2-9B averages 10.68 language-model calls per target and each GUI-Owl-1.5 model averages 34.72, including the initial forward. These timing results are summarized in \cref{tab:recentlatency}.

\paragraph{Accuracy among direct readouts.}
Pro-4B improves mean accuracy over GUI-Swift-4B by 18.03 percentage points, with a higher mean latency of 0.323\,s per target compared with 0.120\,s for GUI-Swift-4B. This accuracy--latency trade-off is reported in \cref{tab:main}.

\paragraph{Lower tail latency.}
Tail latency favors PixelOne-Pro on ScreenSpot, ScreenSpot V2, and OSWorld-G, where each Pro model's P95 latency is below the median latency of UI-Venus-2-9B. On ScreenSpot, for example, Pro-35B-A3B has a P95 latency of 0.450\,s, compared with a median latency of 0.531\,s for UI-Venus-2-9B. Full per-benchmark latency profiles are provided in \cref{tab:arspeed,tab:recentlatency}.

\subsection{What Search Contributes}
\label{sec:distillationeffects}
\begin{table}[!htbp]
\caption{\textbf{Teacher initialization and spatial distillation.} At each scale, students differ only in initialization and distillation. Results are from single training runs.}
\label{tab:ablation}
\centering\small
\setlength{\tabcolsep}{3pt}
\begin{tabular}{@{}ll*{5}{c}@{}}
\toprule
& Student & SS & V2 & Pro & OSW & Avg.\\
\midrule
\multirow{3}{*}{4B}
& Base init, GT & 77.99 & 79.48 & 36.24 & 30.00 & 55.93\\
& Teacher init, GT & 84.91 & 87.89 & 43.77 & 39.02 & 63.90\\
& \quad $+$ distillation & \textbf{87.19} & \textbf{89.78} & \textbf{48.89} & \textbf{46.47} & \textbf{68.08}\\
\midrule
\multirow{3}{*}{35B}
& Base init, GT & 89.78 & 91.98 & 53.19 & 53.92 & 72.22\\
& Teacher init, GT & 90.41 & 92.77 & 57.62 & 59.02 & 74.95\\
& \quad $+$ distillation & \textbf{91.12} & \textbf{93.08} & \textbf{58.57} & \textbf{60.39} & \textbf{75.79}\\
\bottomrule
\end{tabular}
\end{table}

The Top20 ablation in \cref{tab:ablation} examines the contributions of teacher initialization and spatial distillation. Together, they improve mean accuracy over the base-initialized reference by 12.2 percentage points at 4B and 3.6 points at 35B-A3B. All configurations use the training setup detailed in \cref{app:init}, with language LoRA \citep{hu2022lora} and a frozen visual encoder.

\paragraph{Transferring parameters.}
With box supervision alone, teacher-initialized students outperform base-initialized students on every benchmark at both scales. Mean accuracy improves by 8.0 percentage points at 4B and 2.7 at 35B-A3B.

\paragraph{Adding spatial distillation.}
For teacher-initialized students, spatial KD improves mean accuracy by 4.2 points at 4B and 0.8 at 35B-A3B, with gains on all benchmarks at both scales.

\begin{table}[!htbp]
\caption{\textbf{Distilling a distribution rather than a point.} At each scale, students share the single-winner pixel loss and differ only in initialization and teacher supervision. $^\dagger$ marks gains over teacher-point supervision ($p<0.01$, paired sign test).}
\label{tab:hardkd}
\centering\small
\begin{tabular}{lllcccc}
\toprule
& Init & Teacher signal & SS & V2 & Pro & OSW\\
\midrule
\multirow{4}{*}{4B}
& base & none & 81.76 & 83.73 & 37.00 & 30.39\\
& T3 & none & 85.69 & 88.52 & 42.76 & 37.65\\
& T3 & teacher point & 84.67 & 87.03 & 42.06 & 37.25\\
& T3 & distribution & \textbf{87.11}$^\dagger$ & \textbf{89.62}$^\dagger$ & \textbf{44.97}$^\dagger$ & \textbf{40.20}\\
\midrule
\multirow{4}{*}{35B}
& base & none & 88.84 & 90.72 & 50.16 & 49.02\\
& T3 & none & 90.25 & 92.22 & 51.80 & 53.14\\
& T3 & teacher point & 90.33 & 92.37 & 53.57 & 55.29\\
& T3 & distribution & \textbf{90.41} & \textbf{92.85} & \textbf{56.29}$^\dagger$ & \textbf{55.69}\\
\bottomrule
\end{tabular}
\end{table}

\paragraph{Distribution versus point supervision.}
Using the single-winner pixel loss, we separately compare distribution and teacher-point supervision with the same teacher initialization at each scale. As shown in \cref{tab:hardkd}, distribution supervision performs better on every benchmark at both scales, with significant gains on ScreenSpot Pro ($p<0.01$, paired sign test). A single click specifies one location, whereas the distribution retains the teacher's relative preferences across the searched regions.

\subsection{Why the Pixel Objective Matters}
\label{sec:lossdiag}

We examine two design choices: aligning pixel scores with the final click and learning from multiple valid pixels.

\begin{table}[!htbp]
\caption{\textbf{Pixel loss with token selection fixed.} The 35B-A3B students differ only in pixel loss. \Cref{app:winnercontrol} details the setup.}
\label{tab:winner}
\centering\small
\begin{tabular}{lcccc}
\toprule
Pixel loss & SS & V2 & Pro & OSW\\
\midrule
Set likelihood & 89.47 & 91.82 & 46.93 & 49.80\\
One winner (\cref{eq:winner}) & \textbf{90.33} & \textbf{92.77} & \textbf{51.04} & \textbf{50.59}\\
\bottomrule
\end{tabular}
\end{table}

\paragraph{Aligning scores with the winning click.}
With token selection fixed, the winner loss improves accuracy on all four benchmarks in \cref{tab:winner}, gaining 4.1 percentage points on ScreenSpot Pro. For selected tokens containing target pixels, mean in-box probability mass remains similar at 77.4\% for set likelihood and 76.8\% for winner loss. The objective thus converts similar in-box probability mass into more correct clicks through better pixel ranking.

\begin{table}[!htbp]
\caption{\textbf{How many valid pixels to supervise.} Top20 ($\rho=0.2$) achieves the highest mean accuracy among the three settings at both model scales.}
\label{tab:fraction}
\centering\small
\setlength{\tabcolsep}{4pt}
\begin{tabular}{@{}llccccc@{}}
\toprule
& $\rho$ & SS & V2 & Pro & OSW & Avg.\\
\midrule
\multirow{3}{*}{4B}
& one winner & 87.11 & 89.62 & 44.97 & 40.20 & 65.47\\
& \textbf{0.2} & 87.19 & 89.78 & \textbf{48.89} & \textbf{46.47} & \textbf{68.08}\\
& 1.0 & 87.19 & 89.78 & 47.56 & 41.76 & 66.57\\
\midrule
\multirow{3}{*}{35B}
& one winner & 90.41 & 92.85 & 56.29 & 55.69 & 73.81\\
& \textbf{0.2} & \textbf{91.12} & \textbf{93.08} & \textbf{58.57} & \textbf{60.39} & \textbf{75.79}\\
& 1.0 & 90.02 & 92.69 & \textbf{58.57} & 59.02 & 75.07\\
\bottomrule
\end{tabular}
\end{table}

\paragraph{Learning from multiple valid pixels.}
Top20 achieves the highest four-benchmark mean at both scales in \cref{tab:fraction}, improving over one-winner supervision by 2.6 percentage points at 4B and 2.0 at 35B-A3B. Under the same training setup, it supervises the highest-scoring 20\% of target pixels within each token, rather than one or all target pixels. This supports learning from multiple high-scoring valid clicks, motivating the use of Top20 in PixelOne and PixelOne-Pro.

\Needspace{24\baselineskip}
\subsection{Region Selection and Pixel Precision}
\label{sec:mechanism}
\begingroup
\titleformat{\paragraph}[block]{\itshape}{}{0pt}{}
\titlespacing*{\paragraph}{0pt}{6pt}{3pt}

\begin{wrapfigure}{r}{0.51\textwidth}
\centering
\captionsetup{width=\linewidth}
\includegraphics[width=\linewidth]{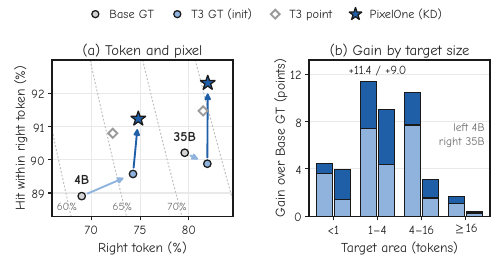}
\caption{\textbf{Search transfer improves grounding.} (a) Target-token coverage and conditional pixel accuracy. (b) Gains over base initialization by target area. Results use single-winner students in \cref{tab:hardkd}.}
\label{fig:mechanism}
\end{wrapfigure}

We analyze the single-winner students in \cref{tab:hardkd}, pooling evaluation targets. Target-token coverage measures how often the selected token contains a valid click, while conditional pixel accuracy measures click accuracy within this subset.

\paragraph{Initialization strengthens region selection.}
Teacher initialization improves target-token coverage at both scales in \cref{fig:mechanism}a. The error breakdown in \cref{app:errors} shows that most of the gain comes from fewer selections far from the target, indicating stronger global localization.

\paragraph{Spatial KD strengthens pixel precision.}
Spatial KD improves conditional pixel accuracy at both scales while maintaining similar target-token coverage. On ScreenSpot Pro at 35B-A3B, within-token pixel errors fall from 251 to 185, while the number of targets with a correct token selection changes only from 1,070 to 1,075. The category results in \cref{tab:categories} show gains for both text and icon targets. Distribution supervision also achieves the highest conditional pixel accuracy among these students and higher target-token coverage than point supervision.

\paragraph{Benefits on compact targets.}
The gains extend to compact targets in \cref{fig:mechanism}b. For boxes with areas equivalent to one to four visual tokens, teacher initialization and spatial KD together improve accuracy over base initialization by 11.4 percentage points at 4B and 9.0 at 35B-A3B.
\par\endgroup

\section{Conclusion}
PixelOne transfers a search teacher's parameters and spatial distributions to a student that predicts clicks directly from the full screenshot in a single forward pass. Its shared decision head selects a visual token, while a lightweight pixel head trained with a click-aligned objective locates the click within that token. Experiments across four benchmarks validate the contributions of spatial distillation and click-aligned pixel learning. PixelOne-Pro-35B-A3B achieves 83.8\% mean accuracy with a mean latency of 0.43\,s per target, while both the 4B and 35B-A3B configurations lie on the accuracy--latency frontier of the evaluated systems. More broadly, PixelOne offers a design principle for efficient GUI grounding: move spatial exploration into training and align supervision with the executed click, enabling richer spatial learning without performing search at inference time.

\bibliographystyle{assets/plainnat}
\bibliography{references}

\clearpage
\section*{Supplementary Material}
\phantomsection
\addcontentsline{toc}{section}{Supplementary Material}
\setcounter{section}{0}
\setcounter{figure}{0}
\setcounter{table}{0}
\setcounter{equation}{0}
\setcounter{proposition}{0}
\renewcommand{\thesection}{S\arabic{section}}
\renewcommand{\thefigure}{S\arabic{figure}}
\renewcommand{\thetable}{S\arabic{table}}
\renewcommand{\theequation}{S\arabic{equation}}
\renewcommand{\theproposition}{S\arabic{proposition}}
\renewcommand{\theHsection}{supp.\arabic{section}}
\renewcommand{\theHfigure}{supp.\arabic{figure}}
\renewcommand{\theHtable}{supp.\arabic{table}}
\renewcommand{\theHequation}{supp.\arabic{equation}}
\providecommand{\theHproposition}{\theproposition}
\renewcommand{\theHproposition}{supp.\arabic{proposition}}
\noindent References without an S prefix refer to the main paper. Citation numbers follow the main paper's bibliography.\par
\makeatletter
\setlength{\@fptop}{0pt}
\setlength{\@fpsep}{10pt}
\setlength{\@fpbot}{0pt plus 1fil}
\makeatother
\Cref{app:data,app:methoddetails,app:mathdetails,app:eval} cover data, method details, mathematical analysis, and evaluation. \Cref{app:results,app:controls,app:cropar,app:newcases} report additional results, controlled studies, local generation, and qualitative visualizations.

\section{Data Construction}
\label{app:data}

\subsection{Training Sets and Source Mixtures}
The controlled set contains 21,557 targets from 1,010 ShowUI screenshots \citep{lin2024showui} covering web and OmniACT desktop interfaces \citep{kapoor2024omniact}. Hyperlinks are capped at 60\%.

\begin{table}[!htbp]
\caption{\textbf{Composition of the continuation datasets.} Entries count grounding targets. Share gives each source's percentage of the 500K set. \Cref{app:data} describes the controlled 21,557-target set.}
\label{tab:appendix_data}
\centering\small
\setlength{\tabcolsep}{3pt}
\begin{tabular}{@{}lrrr@{}}
\toprule
Source & 200K & 500K & Share (\%)\\
\midrule
GroundCUA & 140,500 & 350,500 & 70.10\\
ShowUI-web & 30,000 & 90,000 & 18.00\\
OS-Atlas Linux & 12,000 & 22,000 & 4.40\\
OS-Atlas macOS & 6,000 & 13,000 & 2.60\\
ShowUI-desktop & 1,500 & 1,800 & 0.36\\
OS-Atlas RICO & 8,000 & 19,000 & 3.80\\
OS-Atlas UIBert & 2,000 & 3,700 & 0.74\\
\midrule
Total & 200,000 & 500,000 & 100.00\\
\bottomrule
\end{tabular}
\end{table}

\Cref{tab:appendix_data} summarizes the 200K and 500K continuation mixtures \citep{feizi2026grounding,wu2024osatlas}. They contain 78,530 and 87,517 unique decoded-RGB screenshots, respectively, with one instruction retained per screenshot--box pair.

\subsection{Validation Split and Benchmark Screening}
The 200K and 500K runs share 2,000 held-out targets from 1,808 screenshots, with each source image and its crops assigned to the same split.

Screening against all four benchmarks found no exact decoded-RGB matches or near duplicates. The near-duplicate rule requires dHash distance $\le 2$, grayscale MAE $\le 3$, and log-aspect-ratio difference $\le 0.02$. Both continuation sets use benchmark-filtered sources. The 500K expansion also excludes existing target IDs, duplicate screenshot--box pairs, and validation image groups.

\section{Method Details and Key Settings}
\label{app:methoddetails}
\subsection{Native Input and Pixel Readout}
\label{app:readout}
\paragraph{Input.}
We pad the screenshot on the right and bottom to multiples of stride $s=32$ without resizing. Two identical frames yield one grid of $N=\lceil W/s\rceil\lceil H/s\rceil$ tokens for a $W\times H$ screenshot. The native score scale is fixed at $\alpha=10$.

\paragraph{Pixel head.}
The pixel head uses 128 channels, a $3\times3$ depthwise convolution, and a $128\to128\to1024$ MLP with GELU. Given token-feature grid $v$ and instruction query $q$, it computes
\begin{align}
 v_j&=W_v\operatorname{LN}(h_j),\qquad
 (\gamma,\beta)=W_f\operatorname{LN}(q),\nonumber\\
 \tilde v&=v\odot(1+\tanh\gamma)+\beta,\nonumber\\
 z_j&=\operatorname{MLP}\big[\tilde v_j+
                  \operatorname{DWConv}_{3\times3}(\tilde v)_j\big].
 \label{eq:fine}
\end{align}
The projections $W_v$ and $W_f$ are affine with implicit biases, and the visual states and query share the same LayerNorm parameters.
At inference, the convolution processes the full grid, but the MLP processes only the selected token. The pixel head has 1,138,432 parameters at 4B and 940,800 at 35B-A3B.

With zero-based indices and $C=\lceil W/s\rceil$, pixel $k$ in token $j$ has coordinate
\begin{equation}
 p_{jk}=\big(s(j\bmod C)+(k\bmod s)+\tfrac12,\;
 s\lfloor j/C\rfloor+\lfloor k/s\rfloor+\tfrac12\big).
 \label{eq:coordinate}
\end{equation}
The loss and decoder use the same mask to exclude padding pixels.

\paragraph{Training readouts.}
For GT supervision, we read pixel logits for the union of the box-center token, up to four top-scoring tokens, and up to four top-scoring GT-valid tokens. The GT-valid subset of this union defines $R$ in \cref{eq:gt}. KD also reads teacher-observed tokens without changing $R$.

\subsection{Training the Search Teacher}
\label{app:teacher}
\paragraph{Search states.}
Each local state includes the full image and a crop with a 12\% margin on each side. Each refinement round batches two local states.

\paragraph{Schedule.}
We train the teacher for 500 updates on box-derived states, then 500 on an approximately 1:1 mixture of box-derived and visited states. We collect visited states on 4,096 training targets after the first stage. The final teacher is frozen for student initialization and trace caching.

\paragraph{Labels.}
Row and column supervision uses the box center when it lies inside the crop. Visibility is positive when the crop contains the target box and negative when they do not overlap. We omit visibility loss for partial overlap.

\subsection{Student Initialization and Training}
\label{app:init}
\paragraph{Initialization.}
Base-initialized controls use Qwen weights with the VideoJEV decision head and special-token embeddings. Teacher-initialized controls use T3 weights and a new pixel head.

\paragraph{Controlled training.}
We train rank-16 language LoRA and both heads for 1,500 updates with a frozen visual encoder and causal attention. Matched pairs share pixel-head initialization and data order.

\paragraph{Expanded configurations.}
\label{app:pro}
At each scale, we train both variants independently from the same 21,557-target Top20+KD student, retaining its pixel head.

For \textbf{PixelOne}, we train rank-16 language LoRA and both heads on 200K targets with a frozen visual encoder and causal attention.

For \textbf{PixelOne-Pro}, we train all parameters, including the visual encoder, on 500K targets. Full-attention layers are bidirectional with padding-only masks. Linear attention is unchanged.

Both variants use Top20 and observed-support pixel KD.

\section{Proofs and Distillation Details}
\label{app:mathdetails}
\subsection{Conditional Click Bound}
\label{app:proof}
\begin{proposition}[Conditional click bound]
\label{prop:winner}
For a token $j$ with $A_j\neq\emptyset$ and finite logits, let
$\hat k=\argmax_{k\in V_j}z_{jk}$ and
$k_j^+=\argmax_{k\in A_j}z_{jk}$. Then
\[
 \ind[\hat k\notin A_j]\le\frac{\ell_{jk_j^+}}{\log 2},
\]
where $V_j$ is the decoder's valid-pixel set and $\ell$ is defined in \cref{eq:winner}.
\end{proposition}
\begin{proof}[Proof of \cref{prop:winner}]
If $\hat k\in A_j$, the indicator is zero and the loss is nonnegative. Otherwise, $z_{j\hat k}\ge z_{jk_j^+}$, including ties, so
\[
 \sum_{v\in V_j\setminus A_j}e^{z_{jv}-z_{jk_j^+}}
 \ge e^{z_{j\hat k}-z_{jk_j^+}}\ge1.
\]
Substituting this inequality into the winner loss gives
\[
\begin{aligned}
 \ell_{jk_j^+}
 &=\log\left(1+\sum_{v\in V_j\setminus A_j}
 e^{z_{jv}-z_{jk_j^+}}\right)\\
 &\ge\log(1+1)=\log 2.
\end{aligned}
\]
This proves the bound.
\end{proof}

\subsection{Set Mass and the Pixel Decision}
Suppose $K$ target pixels have logit zero and one non-target pixel has logit $\delta>0$. The non-target pixel wins, but the set loss is
\begin{equation}
 \mathcal L_{\mathrm{set}}=\log(1+e^\delta/K).
\end{equation}
Increasing $K$ lowers this loss without correcting the click. In contrast, every target pixel has winner loss $\ell_{jk}=\log(1+e^\delta)>\log 2$.

Applying \cref{prop:winner} to the selected token gives
\begin{equation}
 \ind[\hat p\notin B]\le
 \begin{cases}
  1, & j^*\notin J_B,\\
  \ell_{j^*k^+_{j^*}}/\log 2, & j^*\in J_B.
 \end{cases}
\end{equation}
\Cref{eq:coarseloss} defines the loss for token selection. The winner loss bounds pixel errors after a target-valid token is selected.

\subsection{Region-KD Normalization}
\label{app:region}
\begingroup
\noindent
\begin{minipage}[t]{0.53\linewidth}
\vspace{0pt}
Region KD compares teacher and student distributions over leaves within each crop. We normalize these distributions separately for each crop, even when crops overlap.

For crop $r$, let $R_{r\ell}$ denote its leaves and $s_{r\ell}$ their teacher log-scores. We discard leaves with no valid pixel centers and normalize the remaining scores as $t_{r\ell}=\softmax(s_{r\cdot}/T)_\ell$, with $T=2$.

For a valid pixel $k\in V_j$, the student's unnormalized mass is
\begin{equation}
 \mu_{jk}=e^{a_j/T}
 \frac{e^{z_{jk}/T}}{\sum_{v\in V_j}e^{z_{jv}/T}}.
\end{equation}
We assign pixels to leaves by their centers using half-open boundaries. We sum their masses within each leaf as
$M_{r\ell}=\sum_{j,k:p_{jk}\in R_{r\ell}}\mu_{jk}$
and normalize within the crop to obtain
$\pi_{r\ell}=M_{r\ell}/\sum_{\ell'}M_{r\ell'}$.
Adding the shared token-level term, including its $T^2$ factor, gives
\begin{equation}
\begin{aligned}
 \mathcal L_{\mathrm{KD}}^{\mathrm{reg}}&=\mathcal L_{\mathrm{KD,coarse}}\\
 &+T^2\sum_r\omega_r\KL\!\left(t_{r\cdot}\,\middle\|\,\pi_{r\cdot}\right).
\end{aligned}
\end{equation}
We set the weights $\omega_r$ proportional to the original parent mass times the crop visibility probability and normalize them over nonempty crops. Region and pixel KD use the same teacher-correct gate. \Cref{tab:regionkd} reports their controlled comparison.
\end{minipage}\hfill
\begin{minipage}[t]{0.44\linewidth}
\begin{minipage}[t]{\linewidth}
\vspace{0pt}
\centering
\captionsetup{width=\linewidth}
\includegraphics[width=\linewidth]{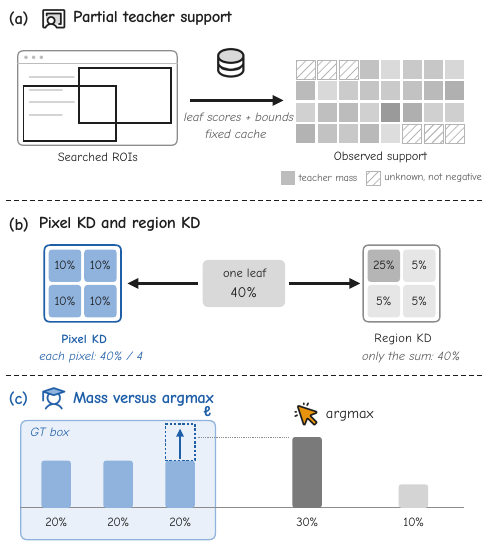}
\captionof{figure}{\textbf{Spatial supervision and the click objective.} Schematic distributions show (a) observed teacher support, (b) region and pixel targets, and (c) a non-target pixel winning despite greater total mass on target pixels.}

\label{fig:supervision}
\end{minipage}
\end{minipage}
\endgroup
\par\allowbreak

\section{Evaluation and Timing Protocols}
\label{app:eval}
\subsection{Benchmarks and Coordinate Conventions}
We evaluate on 1,272 ScreenSpot, 1,272 ScreenSpot V2, 1,581 ScreenSpot Pro, and 510 OSWorld-G rectangular targets. A click is correct if it lies inside the target box in original-image coordinates. Mean accuracy weights the four benchmarks equally, while \cref{app:errors} pools all 4,635 targets. Checkpoints are selected by update count.

\subsection{Base-Model Coordinate Generation}
\label{app:ar}
Qwen3.5-4B and Qwen3.6-35B-A3B use greedy decoding with thinking disabled and KV caching enabled. Screenshots are padded to multiples of 32 without resizing. Both models use the following prompt.
\begin{quote}\small
Instruction: [JSON-quoted instruction]\\
Locate the UI element to click for this instruction in the image, and output its point coordinates in JSON format.
\end{quote}
We map 0--1000 \texttt{point\_2d} coordinates \citep{qwen2025groundingcookbook} from the padded canvas to original pixels. Malformed, ambiguous, out-of-range, and padding predictions count as misses.

\subsection{External Model Configurations}
\label{app:recentbaselines}
The six recent baselines use their released inference policies in Transformers, without action history.

\noindent\textbf{GUI-Decisions-2B} uses its released slot readout with an input cap of approximately 1 megapixel.

\noindent\textbf{GUI-Decisions-31B} uses the BF16 merged Gemma-4-31B E3 release with a budget of 280 visual tokens.

\noindent\textbf{GUI-Swift-4B} uses the Desktop131 adapter on GUI-Owl-1.5-4B with 131 slots in one causal forward and an adaptive image budget. Actions without coordinates count as misses.

\noindent\textbf{GUI-Owl-1.5-4B/8B} both use Instruct checkpoints, the official grounding prompt and parser, and a cap of 9,800 visual tokens.

\noindent\textbf{UI-Venus-2-9B} uses the official direct-coordinate prompt and parser with thinking disabled and a cap of 12,845,056 pixels. Linear attention uses native FLA and causal-convolution kernels.

\paragraph{Training counts.}
Counts in \cref{tab:main} cover reported adaptation stages, excluding earlier or inherited training. The 31B Decisions stage includes natural images. Full training totals for GUI-Owl and UI-Venus-2 remain unverified.

\subsection{Timing and Aggregation}
\label{app:timing}
\paragraph{Setup.}
We measure latency on an H200 at batch size 1. The same 128 targets per benchmark are timed three times after warmup, giving 384 measurements per benchmark.

\paragraph{Scope.}
Timing starts from a decoded RGB screenshot and includes preprocessing, visual encoding, inference, and click extraction. AR timing includes full response generation and parsing. Image-file reading, model loading, and warmup are excluded.

\paragraph{Aggregation.}
Mean, P50, and P95 are computed per benchmark. Overall latency is the mean of the four benchmark means. Speedups are ratios of these averages. For cascades, we sum stage times paired by target and repetition, excluding model loading and switching.

% \FloatBarrier
\section{Additional Grounding Results}
\label{app:results}
\begin{figure}[!htbp]
\centering
\includegraphics[width=\textwidth]{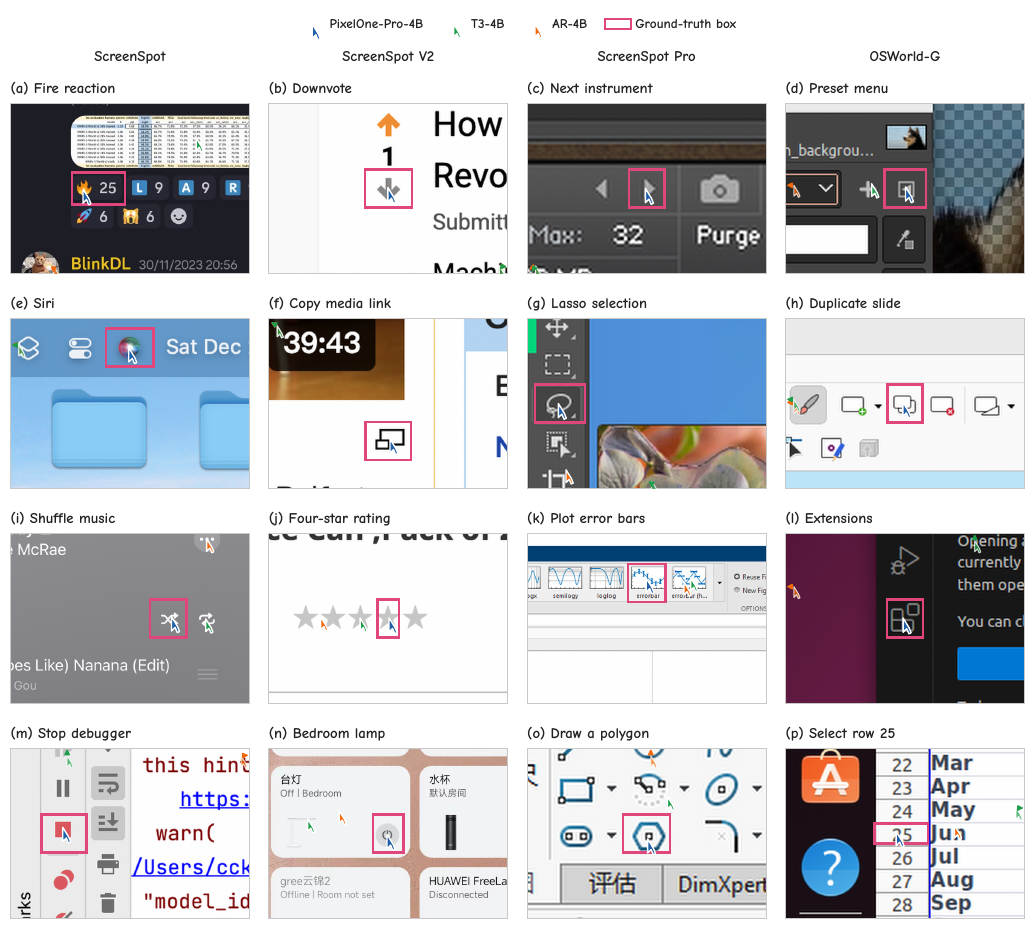}
\caption{\textbf{Single-forward grounding across four benchmarks.} In all sixteen selected cases, four per benchmark, PixelOne-Pro-4B succeeds and same-scale T3 and AR fail. Columns show ScreenSpot, V2, Pro, and OSWorld-G. Crops are for display only. Boxes mark targets, and border arrows indicate predictions outside the crop.}
\label{fig:qualgallery}
\end{figure}
\subsection{Latency Across Benchmarks}
\begin{table}[!htbp]
\captionsetup{width=\textwidth}
\centering
\begin{minipage}[t]{0.485\textwidth}
\vspace{0pt}
\begingroup
\captionsetup{width=\linewidth}
\captionof{table}{\textbf{Latency of direct prediction and coordinate generation.} Mean, P50, and P95 are in seconds per target, and Calls is the average number of model calls. PixelOne and PixelOne-Pro use the 200K and 500K continuations, respectively. See \cref{app:timing} for measurement details.}

\label{tab:arspeed}
\centering\small
\setlength{\tabcolsep}{2pt}
\begin{tabularx}{\linewidth}{@{}>{\raggedright\arraybackslash}Xlrrrr@{}}
\toprule
Model & Set & Mean & P50 & P95 & Calls\\
\midrule
\multirow{4}{*}{\shortstack[l]{PixelOne-Pro\\4B, 500K}} & SS & \underline{0.248} & \underline{0.257} & \underline{0.343} & 1\\
 & V2 & \underline{0.249} & \underline{0.258} & \underline{0.349} & 1\\
 & Pro & \underline{0.642} & \underline{0.389} & \underline{1.871} & 1\\
 & OSW & \textbf{0.153} & \textbf{0.170} & \underline{0.191} & 1\\
\midrule
\multirow{4}{*}{\shortstack[l]{PixelOne\\4B, 200K}} & SS & \textbf{0.245} & \textbf{0.254} & \textbf{0.338} & 1\\
 & V2 & \textbf{0.244} & \textbf{0.254} & \textbf{0.339} & 1\\
 & Pro & \textbf{0.623} & \textbf{0.369} & \textbf{1.819} & 1\\
 & OSW & \underline{0.158} & \underline{0.172} & \textbf{0.188} & 1\\
\midrule
\multirow{4}{*}{\shortstack[l]{Qwen3.5-4B\\AR}} & SS & 1.502 & 1.282 & 3.188 & 46\\
 & V2 & 1.473 & 1.320 & 2.768 & 45\\
 & Pro & 1.698 & 1.480 & 2.886 & 39\\
 & OSW & 1.329 & 1.208 & 2.539 & 43\\
\midrule
\multirow{4}{*}{\shortstack[l]{PixelOne-Pro\\35B-A3B, 500K}} & SS & \underline{0.328} & \underline{0.337} & \underline{0.450} & 1\\
 & V2 & \underline{0.332} & \underline{0.342} & \underline{0.457} & 1\\
 & Pro & \underline{0.849} & \underline{0.484} & \underline{2.557} & 1\\
 & OSW & \underline{0.213} & \textbf{0.220} & \underline{0.264} & 1\\
\midrule
\multirow{4}{*}{\shortstack[l]{PixelOne\\35B-A3B, 200K}} & SS & \textbf{0.321} & \textbf{0.330} & \textbf{0.434} & 1\\
 & V2 & \textbf{0.320} & \textbf{0.329} & \textbf{0.433} & 1\\
 & Pro & \textbf{0.814} & \textbf{0.470} & \textbf{2.446} & 1\\
 & OSW & \textbf{0.212} & \underline{0.230} & \textbf{0.244} & 1\\
\midrule
\multirow{4}{*}{\shortstack[l]{Qwen3.6-35B-A3B\\AR}} & SS & 2.138 & 2.132 & 2.472 & 36\\
 & V2 & 2.196 & 2.156 & 2.581 & 37\\
 & Pro & 2.720 & 2.454 & 4.284 & 37\\
 & OSW & 2.210 & 2.181 & 2.650 & 39\\
\bottomrule
\end{tabularx}
\par
\endgroup

\end{minipage}\hfill
\begin{minipage}[t]{0.485\textwidth}
\vspace{0pt}
\begingroup
\captionsetup{width=\linewidth}
\captionof{table}{\textbf{Latency of recent external models.} Latency is in seconds per target, based on 384 measurements per benchmark following \cref{app:timing}. Calls is the mean number of language-model calls, including the initial forward and decoding steps. All models encode the screenshot once.}
\label{tab:recentlatency}
\centering\small
\setlength{\tabcolsep}{2pt}
\begin{tabularx}{\linewidth}{@{}>{\raggedright\arraybackslash}Xlrrrr@{}}
\toprule
Model & Set & Mean & P50 & P95 & Calls\\
\midrule
\multirow{4}{*}{\shortstack[l]{GUI-Decisions\\2B}} & SS & \textbf{0.079} & \textbf{0.084} & \textbf{0.092} & 1\\
 & V2 & \textbf{0.079} & \textbf{0.082} & \textbf{0.093} & 1\\
 & Pro & \textbf{0.108} & \textbf{0.093} & \textbf{0.185} & 1\\
 & OSW & \textbf{0.067} & \textbf{0.074} & \textbf{0.077} & 1\\
\midrule
\multirow{4}{*}{\shortstack[l]{GUI-Decisions\\31B}} & SS & \textbf{0.137} & \textbf{0.135} & \textbf{0.159} & 1\\
 & V2 & \textbf{0.130} & \textbf{0.130} & \textbf{0.144} & 1\\
 & Pro & \textbf{0.160} & \textbf{0.140} & \textbf{0.265} & 1\\
 & OSW & \textbf{0.124} & \textbf{0.124} & \textbf{0.129} & 1\\
\midrule
\multirow{4}{*}{\shortstack[l]{GUI-Swift\\4B}} & SS & \underline{0.116} & \underline{0.118} & \underline{0.138} & 1\\
 & V2 & \underline{0.119} & \underline{0.120} & \underline{0.148} & 1\\
 & Pro & \underline{0.145} & \underline{0.129} & \underline{0.253} & 1\\
 & OSW & \underline{0.100} & \underline{0.101} & \underline{0.113} & 1\\
\midrule
\multirow{4}{*}{\shortstack[l]{GUI-Owl-1.5\\4B}} & SS & 0.953 & 0.973 & 1.064 & 34.69\\
 & V2 & 0.954 & 0.968 & 1.061 & 34.77\\
 & Pro & 1.263 & 1.078 & 1.865 & 34.60\\
 & OSW & 0.846 & 0.855 & 0.902 & 34.83\\
\midrule
\multirow{4}{*}{\shortstack[l]{GUI-Owl-1.5\\8B}} & SS & 1.019 & 1.042 & 1.149 & 34.70\\
 & V2 & 1.020 & 1.033 & 1.150 & 34.79\\
 & Pro & 1.429 & 1.180 & 2.245 & 34.60\\
 & OSW & 0.885 & 0.905 & 0.933 & 34.80\\
\midrule
\multirow{4}{*}{\shortstack[l]{UI-Venus-2\\9B}} & SS & \underline{0.508} & \underline{0.531} & \underline{0.615} & 10.62\\
 & V2 & \underline{0.514} & \underline{0.527} & \underline{0.617} & 10.77\\
 & Pro & \underline{0.902} & \underline{0.633} & \underline{2.105} & 10.56\\
 & OSW & \underline{0.403} & \underline{0.423} & \underline{0.433} & 10.77\\
\bottomrule
\end{tabularx}
\par
\endgroup

\end{minipage}
\end{table}
\Cref{tab:arspeed,tab:recentlatency} report per-benchmark latency and language-model call counts under the protocol in \cref{app:timing}.

\subsection{Search and Direct Prediction}
\label{app:searchcomparison}
\begin{table}[!htbp]
\captionsetup{width=\textwidth}
\caption{\textbf{Search and direct prediction at both scales.} We report accuracy (\%) on complete benchmarks and mean latency on an H200 with batch size 1. Students start from the 21,557-target controls and continue training on 200K or 500K targets. $P$ denotes a prefill call. T3 uses five visual encodings, and each student uses one. Evaluation and timing follow \cref{app:eval,app:timing}.}
\label{tab:searchcomparison}
\centering\small
\setlength{\tabcolsep}{5pt}
\begin{tabular}{@{}llclcccccc@{}}
\toprule
Scale & System & Calls & GUI data & SS & V2 & Pro & OSW & Avg. & s/target\\
\midrule
4B & Teacher T1 & $1P$ & 21.6K tgt. & 19.58 & 19.34 & 1.20 & 4.51 & 11.16 & 0.382\\
& Teacher T3 & $3P$ & 21.6K tgt. & 87.26 & 90.57 & 55.22 & \underline{57.65} & 72.67 & 2.448\\
& PixelOne & $1P$ & 200K tgt. & \textbf{90.64} & \textbf{93.16} & \underline{55.34} & 57.45 & \underline{74.15} & \textbf{0.317}\\
& PixelOne-Pro & $1P$ & 500K tgt. & \underline{90.33} & \underline{92.85} & \textbf{68.12} & \textbf{71.18} & \textbf{80.62} & \underline{0.323}\\
\midrule
35B-A3B & Teacher T1 & $1P$ & 21.6K tgt. & 18.87 & 19.10 & 1.27 & 5.10 & 11.08 & 0.491\\
& Teacher T3 & $3P$ & 21.6K tgt. & 90.80 & \underline{93.79} & \underline{67.36} & 68.24 & \underline{80.05} & 2.899\\
& PixelOne & $1P$ & 200K tgt. & \underline{91.04} & 93.32 & 63.50 & \underline{68.82} & 79.17 & \textbf{0.417}\\
& PixelOne-Pro & $1P$ & 500K tgt. & \textbf{91.67} & \textbf{94.50} & \textbf{72.04} & \textbf{76.86} & \textbf{83.77} & \underline{0.431}\\
\bottomrule
\end{tabular}
\end{table}

\Cref{tab:searchcomparison} compares search teachers with the 200K and 500K students. Pro-35B-A3B improves on T3 across all four benchmarks, increasing mean accuracy by 3.72 percentage points with a $6.73\times$ speedup.

\subsection{Qualitative Examples}
\label{app:qual}
\Cref{fig:qualgallery} shows Pro-4B model comparisons, while \cref{fig:qualitative} pairs Pro-35B-A3B predictions with the teacher's search trajectories.

\subsection{Published Reference Results}
\label{app:external}
\begin{table}[!htbp]
\captionsetup{width=\textwidth}
\caption{\textbf{Published grounding results and local evaluations.} Accuracy (\%) is rounded to one decimal. Published scores retain their original protocols. GUI-Actor and our models are evaluated locally. GUI data uses the reported units, with screenshot counts in parentheses. $P$, $D$, and $R$ denote prefill, autoregressive decoding, and reranking.}

\label{tab:external}
\centering\footnotesize
\setlength{\tabcolsep}{2pt}
\renewcommand{\arraystretch}{1.08}
\begin{tabular}{@{}>{\raggedright\arraybackslash}p{0.315\textwidth}clrrrrl>{\raggedright\arraybackslash}p{0.11\textwidth}@{}}
\toprule
Method and backbone & Calls & GUI data & SS & V2 & Pro & OSW & Protocol & Source\\
\midrule
\multicolumn{9}{@{}l}{\textit{Single-pass coordinate-free readout}}\\
GUI-Actor-3B {\scriptsize(Qwen2.5-VL-3B)} & $1P$ & 9.6M elem. (1M) & 88.3 & 90.6 & 43.2 & 44.1 & R510 & Ours, official\\
GUI-Actor-7B {\scriptsize(Qwen2.5-VL-7B)} & $1P$ & 9.6M elem. (1M) & 89.9 & 92.1 & 46.4 & 46.1 & R510 & Ours, official\\
GUI-AIMA-3B {\scriptsize(Qwen2.5-VL-3B)} & $1P$ & 509K samp. (101K) & -- & 92.1 & 53.8 & 62.8 & Ref/U & \citep{zhou2025guiaima}, T1--3\\
V2P-3B {\scriptsize(Qwen2.5-VL-3B)} & $1P$ & 9.6M elem. (1M) & -- & 91.4 & 48.5 & 48.8 & U & \citep{chen2025v2p}, T1\\
V2P-7B {\scriptsize(Qwen2.5-VL-7B)} & $1P$ & 9.6M elem. (1M) & -- & 92.4 & 52.5 & 52.5 & U & \citep{chen2025v2p}, T1\\
Seg2Click-4B {\scriptsize(Qwen3.5-4B)} & $1P$ & 1.26M rec. (${\sim}$1M) & 91.6 & 93.1 & 50.2 & -- & -- & \citep{wang2026seg2click}, T1--3\\
Seg2Click-9B {\scriptsize(Qwen3.5-9B)} & $1P$ & 1.26M rec. (${\sim}$1M) & 92.1 & 94.1 & 55.2 & -- & -- & \citep{wang2026seg2click}, T1--3\\
\textbf{PixelOne-4B} {\scriptsize(VideoJEV-4B)} & $1P$ & 21.6K\,$\to$\,200K tgt. & 90.6 & 93.2 & 55.3 & 57.5 & R510 & Ours, \cref{tab:main}\\
\textbf{PixelOne-35B-A3B} {\scriptsize(VideoJEV-35B-A3B)} & $1P$ & 21.6K\,$\to$\,200K tgt. & 91.0 & 93.3 & 63.5 & 68.8 & R510 & Ours, \cref{tab:main}\\
\textbf{PixelOne-Pro-4B} {\scriptsize(VideoJEV-4B)} & $1P$ & 21.6K\,$\to$\,500K tgt. & 90.3 & 92.8 & 68.1 & 71.2 & R510 & Ours, \cref{tab:main}\\
\textbf{PixelOne-Pro-35B-A3B} {\scriptsize(VideoJEV-35B-A3B)} & $1P$ & 21.6K\,$\to$\,500K tgt. & 91.7 & 94.5 & 72.0 & 76.9 & R510 & Ours, \cref{tab:main}\\
\midrule
\multicolumn{9}{@{}l}{\textit{Autoregressive coordinate generation}}\\
UI-TARS-7B {\scriptsize(Qwen2-VL-7B)} & $1P+D$ & 18.4M elem.$^\ast$ & 89.5 & 91.6 & 35.7 & -- & -- & \citep{qin2025uitars}, T4--6\\
UI-TARS-72B {\scriptsize(Qwen2-VL-72B)} & $1P+D$ & 18.4M elem.$^\ast$ & 88.4 & 90.3 & 38.1 & -- & -- & \citep{qin2025uitars}, T4--6\\
Jedi-3B {\scriptsize(Qwen2.5-VL-3B)} & $1P+D$ & 4M ex. & -- & 88.6 & 36.1 & 50.9 & F564 & \citep{xie2025jedi}, T2, 3, 5\\
Jedi-7B {\scriptsize(Qwen2.5-VL-7B)} & $1P+D$ & 4M ex. & -- & 91.7 & 39.5 & 54.1 & F564 & \citep{xie2025jedi}, T2, 3, 5\\
InnerZoom-4B {\scriptsize(Qwen3-VL-4B)} & $1P+D$ & 283K\,+\,100K samp.$^\S$ & -- & 95.2 & 66.2 & 64.7 & O/U & \citep{liu2026innerzoom}, T1, 2\\
\textbf{Qwen3.5-4B} & $1P+D$ & none & 87.8 & 90.6 & 55.0 & 56.7 & R510 & Ours, \cref{tab:main}\\
\textbf{Qwen3.6-35B-A3B} & $1P+D$ & none & 91.5 & 93.7 & 60.0 & 76.5 & R510 & Ours, \cref{tab:main}\\
\midrule
\multicolumn{9}{@{}l}{\textit{Localization followed by zoom or another refinement pass}}\\
GUI-AIMA-3B $+$ zoom {\scriptsize(Qwen2.5-VL-3B)} & $2P$ & 509K samp. (101K) & -- & -- & 61.5 & 68.1 & Ref/U & \citep{zhou2025guiaima}, T1, 3\\
RankGround-8B {\scriptsize(UI-Venus-1.5-8B)} & $R+1P+D$ & 6.8K inst.$^\ddagger$ & -- & 95.4 & 71.0 & 78.0 & O/U & \citep{fan2026rankground}, T2, 8\\
Re-Prefill-8B {\scriptsize(GUI-Owl-1.5-8B)} & varies$^\dagger$ & none & -- & 94.6 & 80.1 & 68.8 & O/U & \citep{lin2026prefill}, T1--3\\
\textbf{PixelOne $+$ AR-4B} {\scriptsize(1024)} & $2P+D$ & 21.6K\,$\to$\,200K tgt. & 88.8 & 91.0 & 65.5 & 67.3 & R510 & Ours, \cref{tab:cropar}\\
\textbf{PixelOne $+$ AR-4B} {\scriptsize(1536)} & $2P+D$ & 21.6K\,$\to$\,200K tgt. & 88.9 & 91.5 & 64.5 & 67.6 & R510 & Ours, \cref{tab:cropar}\\
\textbf{PixelOne-Pro $+$ AR-4B} {\scriptsize(1024)} & $2P+D$ & 21.6K\,$\to$\,500K tgt. & 87.5 & 90.0 & 67.0 & 68.0 & R510 & Ours, \cref{tab:croparspeed}\\
\textbf{PixelOne-Pro $+$ AR-4B} {\scriptsize(1536)} & $2P+D$ & 21.6K\,$\to$\,500K tgt. & 87.7 & 90.4 & 64.8 & 67.8 & R510 & Ours, \cref{tab:croparspeed}\\
\textbf{PixelOne $+$ AR-35B-A3B} {\scriptsize(1024)} & $2P+D$ & 21.6K\,$\to$\,200K tgt. & 91.8 & 93.6 & 73.2 & 78.6 & R510 & Ours, \cref{tab:cropar}\\
\textbf{PixelOne $+$ AR-35B-A3B} {\scriptsize(1536)} & $2P+D$ & 21.6K\,$\to$\,200K tgt. & 92.8 & 94.9 & 74.3 & 79.8 & R510 & Ours, \cref{tab:cropar}\\
\textbf{PixelOne-Pro $+$ AR-35B-A3B} {\scriptsize(1024)} & $2P+D$ & 21.6K\,$\to$\,500K tgt. & 91.7 & 93.9 & 74.6 & 80.2 & R510 & Ours, \cref{tab:croparspeed}\\
\textbf{PixelOne-Pro $+$ AR-35B-A3B} {\scriptsize(1536)} & $2P+D$ & 21.6K\,$\to$\,500K tgt. & 92.3 & 95.0 & 75.0 & 81.2 & R510 & Ours, \cref{tab:croparspeed}\\
\bottomrule
\end{tabular}
\par\smallskip
\parbox{\textwidth}{\footnotesize
\textit{Protocols.} R510 contains 510 rectangular targets. F564 contains 564 items, including refusal cases. O/U uses the original annotations, with counts and refusal handling unspecified. Ref/U uses refined annotations with unspecified counts. U denotes an unspecified protocol.\par
\textit{Sources and units.} T refers to a table in the cited paper, while -- marks an unreported value. Unit abbreviations are elem. for elements, rec. for records, samp. for samples, ex. for examples, inst. for instances, and tgt. for targets. \Cref{tab:appendix_data} lists our continuation mixtures.\par
\textit{Configurations.} Each cascade uses its corresponding base AR model with $2\times$ crop upscaling. Parentheses give window sizes in pixels.
$^\ast$UI-TARS also uses an undisclosed amount of in-house data.
$^\ddagger$RankGround reports the training count only for its 2B reranker, not its released 8B grounder.
$^\S$The SFT and RL data may overlap.
$^\dagger$Re-Prefill uses $2P+D$ on V2. Total calls with external zoom are not reported for Pro or OSWorld-G.}

\end{table}

\Cref{tab:external} lists published and locally evaluated results with their protocols. \Cref{tab:main} compares accuracy and latency on the same hardware.

% \FloatBarrier

\section{Controlled Design Studies}
\label{app:controls}
\subsection{Teacher Adaptation and Search Resolution}
\Cref{tab:teacher,tab:rounds,tab:coverage} examine teacher adaptation, search depth, and cell-center resolution, respectively.

\begin{table}[!htbp]
\caption{\textbf{Adapting the 35B search teacher.} We report accuracy (\%) for a fixed three-round search with released weights, after 500 box-supervised updates, and after 1,000 total updates.}
\label{tab:teacher}
\centering\small
\setlength{\tabcolsep}{5pt}
\begin{tabular}{lcccc}
\toprule
Teacher & ScreenSpot & V2 & Pro & OSWorld-G\\
\midrule
Released weights & 83.02 & 86.24 & 50.16 & 52.75\\
Supervised 500 & \underline{90.57} & \textbf{93.95} & \underline{65.97} & \textbf{69.02}\\
Final 1,000 & \textbf{90.80} & \underline{93.79} & \textbf{67.36} & \underline{68.24}\\
\bottomrule
\end{tabular}
\end{table}
\begin{table}[!htbp]
\caption{\textbf{Accuracy after successive search rounds.} T$k$ stops after round $k$ of the $4/8/8$ search with two retained candidates. PixelOne uses the 200K continuation. Avg.\ weights the four benchmarks equally.}
\label{tab:rounds}
\centering\small
\setlength{\tabcolsep}{3.6pt}
\begin{tabular}{lllccccc}
\toprule
& Weights & & SS & V2 & Pro & OSW & Avg.\\
\midrule
\multirow{7}{*}{4B}
& \multirow{3}{*}{Released} & T1 & 16.43 & 16.19 & 0.70 & 3.33 & 9.16\\
& & T2 & 70.52 & 73.11 & 7.59 & 20.59 & 42.95\\
& & T3 & 74.21 & 77.36 & 26.82 & 32.75 & 52.78\\
\cmidrule(l){2-8}
& \multirow{3}{*}{Trained} & T1 & 19.58 & 19.34 & 1.20 & 4.51 & 11.16\\
& & T2 & 83.73 & 86.56 & 17.39 & 36.86 & 56.13\\
& & T3 & \underline{87.26} & \underline{90.57} & \underline{55.22} & \textbf{57.65} & \underline{72.67}\\
\cmidrule(l){2-8}
& \multicolumn{2}{l}{PixelOne} & \textbf{90.64} & \textbf{93.16} & \textbf{55.34} & \underline{57.45} & \textbf{74.15}\\
\midrule
\multirow{7}{*}{35B}
& \multirow{3}{*}{Released} & T1 & 18.32 & 18.40 & 1.20 & 3.92 & 10.46\\
& & T2 & 80.66 & 83.10 & 13.98 & 30.78 & 52.13\\
& & T3 & 83.02 & 86.24 & 50.16 & 52.75 & 68.04\\
\cmidrule(l){2-8}
& \multirow{3}{*}{Trained} & T1 & 18.87 & 19.10 & 1.27 & 5.10 & 11.08\\
& & T2 & 86.87 & 89.62 & 19.17 & 41.96 & 59.40\\
& & T3 & \underline{90.80} & \textbf{93.79} & \textbf{67.36} & \underline{68.24} & \textbf{80.05}\\
\cmidrule(l){2-8}
& \multicolumn{2}{l}{PixelOne} & \textbf{91.04} & \underline{93.32} & \underline{63.50} & \textbf{68.82} & \underline{79.17}\\
\bottomrule
\end{tabular}
\end{table}

\begin{table}[!htbp]
\caption{\textbf{Cell selection versus center-click accuracy on ScreenSpot Pro.} Right cell measures whether the selected cell of the $4\times4$ grid contains the target-box center. Cell-center click measures whether the selected cell's center lies inside the box. Even the best choice among all sixteen centers can hit only 1.39\% of targets.}
\label{tab:coverage}
\centering\small
\begin{tabular}{llcc}
\toprule
& Weights & Right cell & Cell-center click\\
\midrule
\multirow{2}{*}{4B} & Released & \underline{34.28} & \underline{0.70}\\
& Trained & \textbf{67.99} & \textbf{1.20}\\
\midrule
\multirow{2}{*}{35B} & Released & \underline{52.25} & \underline{1.20}\\
& Trained & \textbf{68.75} & \textbf{1.27}\\
\bottomrule
\end{tabular}
\end{table}

\subsection{Pixel and Region Distribution Supervision}
\begin{table}[!htbp]
\caption{\textbf{Pixel versus region distillation.} We report accuracy (\%) for teacher-initialized students using the single-winner pixel loss. All settings use the same 21,557 targets and training budget.}
\label{tab:regionkd}
\centering\small
\setlength{\tabcolsep}{3pt}
\begin{tabular}{@{}llrrrr@{}}
\toprule
Scale & Supervision & SS & V2 & Pro & OSW\\
\midrule
\multirow{3}{*}{4B}
& GT & 85.69 & 88.52 & \underline{42.76} & 37.65\\
& GT + region KD & \underline{86.95} & \underline{89.54} & \textbf{44.97} & \underline{39.80}\\
& GT + pixel KD & \textbf{87.11} & \textbf{89.62} & \textbf{44.97} & \textbf{40.20}\\
\midrule
\multirow{3}{*}{35B}
& GT & 90.25 & 92.22 & 51.80 & 53.14\\
& GT + region KD & \textbf{90.80} & \textbf{93.00} & \underline{54.90} & \underline{55.49}\\
& GT + pixel KD & \underline{90.41} & \underline{92.85} & \textbf{56.29} & \textbf{55.69}\\
\bottomrule
\end{tabular}
\end{table}

All students in \cref{tab:regionkd} use teacher initialization, the single-winner loss, and 48,000 training exposures. Box supervision is identical, and the two KD variants share the teacher-correct gate.

\subsection{Pixel Objectives Under Matched Conditions}
\label{app:winnercontrol}
Both branches in \cref{tab:winner} start from the same 1,400-update checkpoint. They train only the pixel head for 100 updates using the same 3,200 exposures and KD targets. Their token logits are bit-identical on all 4,635 evaluation targets. Mean in-box mass is computed over the same 1,038 target-valid selections on ScreenSpot Pro.

\paragraph{Supervising several valid pixels.}
\label{app:fraction}
The settings in \cref{tab:fraction} share 1,500 updates, teacher initialization, spatial KD, and data order. They differ only in supervising one winner, the top 20\% of valid pixels, or all valid pixels. At 4B, Top20 improves accuracy over one-winner supervision on Pro and OSWorld-G. Paired sign tests give $p<10^{-4}$ and $p<10^{-3}$, respectively.

\subsection{Region Selection and Pixel Precision}
\label{app:errors}
\begingroup
\noindent
\begin{minipage}[t]{0.49\linewidth}
\vspace{0pt}
A wrong token is adjacent if it lies within one grid step of a GT-valid token, and distant otherwise. Pixel errors occur when the selected token is GT-valid but the click misses. Unlike the pooled rates in \cref{fig:errors}, conditional pixel accuracy in \cref{fig:mechanism} uses only GT-valid token selections.

\Cref{tab:categories} reports 35B-A3B results on ScreenSpot Pro text and icon targets.
\par\smallskip
\begingroup
\captionsetup{width=\linewidth}
\captionof{table}{\textbf{Spatial KD on text and icon targets.} Entries give correct clicks and accuracy on ScreenSpot Pro. The 35B students in \cref{tab:hardkd} use teacher initialization and the single-winner loss.}
\label{tab:categories}
\centering\small
\begin{tabular}{lrr}
\toprule
Student & Text (977) & Icon (604)\\
\midrule
GT-only & \underline{663 (67.86\%)} & \underline{156 (25.83\%)}\\
GT + pixel KD & \textbf{714 (73.08\%)} & \textbf{176 (29.14\%)}\\
\bottomrule
\end{tabular}
\par
\endgroup

\end{minipage}\hfill
\begin{minipage}[t]{0.48\linewidth}
\begin{minipage}[t]{\linewidth}
\vspace{0pt}
\centering
\captionsetup{width=\linewidth}
\includegraphics[width=\linewidth]{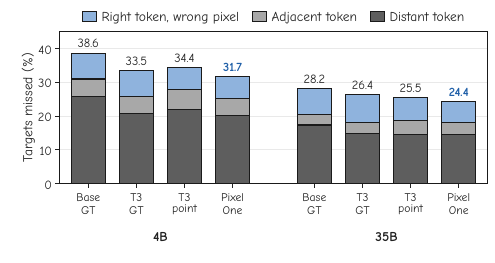}
\captionof{figure}{\textbf{Error breakdown for the single-winner controls.} Models are from \cref{tab:hardkd}. Rates use all 4,635 evaluation targets. Bar labels give total miss rates.}
\label{fig:errors}
\end{minipage}
\end{minipage}
\endgroup

\section{Local Generation from Pixel Predictions}
\label{app:cropar}
\label{sec:fastslow}
\subsection{Crop Windows and Coordinate Mapping}
For $w\in\{1024,1536\}$, we center a window on the PixelOne or PixelOne-Pro click and shift it to fit inside the screenshot. If an image dimension is smaller than $w$, the window spans that dimension. We upscale the crop $2\times$ with LANCZOS and run the corresponding base AR model with the settings in \cref{app:ar}. We map the decoded pixel coordinate back by halving it and adding the crop origin.

\subsection{Accuracy and Paired-Stage Latency}
\begin{table}[!htbp]
\captionsetup{width=\textwidth}
\caption{\textbf{Accuracy and latency of local AR refinement.} Avg.\ and Acc.\ are four-benchmark mean latency (s) and accuracy (\%). Pro denotes ScreenSpot Pro. Data counts continuation targets, and Window is in original pixels. Timings follow \cref{app:timing}.}
\label{tab:croparspeed}
\centering\small
\setlength{\tabcolsep}{4pt}
\begin{tabular}{@{}llccrrrr@{}}
\toprule
& System & Data & Window & Avg.\ (s) & Pro (s) & Pro P95 (s) & Acc.\\
\midrule
\multirow{7}{*}{4B}
& PixelOne & 200K & -- & \textbf{0.317} & \textbf{0.623} & \textbf{1.819} & 74.15\\
& AR & -- & -- & \underline{1.501} & \underline{1.698} & \underline{2.886} & 72.54\\
& T3 & -- & -- & 2.448 & 2.961 & 4.873 & 72.67\\
& PixelOne $\to$ AR & 200K & 1024 & 1.860 & 2.181 & 3.362 & \underline{78.14}\\
& PixelOne $\to$ AR & 200K & 1536 & 2.386 & 3.029 & 4.129 & \textbf{78.15}\\
& PixelOne-Pro $\to$ AR & 500K & 1024 & 1.869 & 2.175 & 3.416 & 78.13\\
& PixelOne-Pro $\to$ AR & 500K & 1536 & 2.293 & 2.764 & 4.084 & 77.69\\
\midrule
\multirow{7}{*}{35B-A3B}
& PixelOne & 200K & -- & \textbf{0.417} & \textbf{0.814} & \textbf{2.446} & 79.17\\
& AR & -- & -- & \underline{2.316} & \underline{2.720} & \underline{4.284} & 80.41\\
& T3 & -- & -- & 2.899 & 3.507 & 5.844 & 80.05\\
& PixelOne $\to$ AR & 200K & 1024 & 2.795 & 3.251 & 4.790 & 84.32\\
& PixelOne $\to$ AR & 200K & 1536 & 3.332 & 4.020 & 5.715 & \underline{85.46}\\
& PixelOne-Pro $\to$ AR & 500K & 1024 & 2.814 & 3.285 & 4.899 & 85.13\\
& PixelOne-Pro $\to$ AR & 500K & 1536 & 3.309 & 4.002 & 5.665 & \textbf{85.86}\\
\bottomrule
\end{tabular}
\end{table}
\Cref{tab:external,tab:croparspeed} report accuracy and latency, while \cref{tab:cropar} adds 200K window diagnostics. Relative to direct prediction in \cref{tab:main}, both windows improve mean accuracy for the 200K models and Pro-35B-A3B, but lower it for Pro-4B. All cascades increase latency.
\begin{table}[!htbp]
\captionsetup{width=\textwidth}
\caption{\textbf{Window diagnostics for the 200K PixelOne models.} Accuracy and recall are percentages. Recall measures how often the target-box center lies inside the window, and AR$\,|$in gives accuracy on those targets. Parse counts malformed outputs, which are treated as misses. Visual token counts are given for the crop and full screenshot.}
\label{tab:cropar}
\centering\small
\begin{tabular}{llcrrrrrrc}
\toprule
& Benchmark & Window & PixelOne & Full AR & Crop AR & Recall & AR\,$|$\,in & Parse & Vis.\ tokens\\
\midrule
\multirow{8}{*}{35B}
& ScreenSpot & 1024 & 91.04 & 91.51 & \underline{91.82} & 96.38 & 94.54 & 10 & 3,840 / 2,734\\
& ScreenSpot & 1536 & 91.04 & 91.51 & \textbf{92.85} & 98.27 & 94.48 & 8 & 7,001 / 2,734\\
& V2 & 1024 & 93.32 & 93.71 & \underline{93.63} & 97.25 & 95.80 & 3 & 3,846 / 2,715\\
& V2 & 1536 & 93.32 & 93.71 & \textbf{94.89} & 98.58 & 96.25 & 2 & 6,987 / 2,715\\
& Pro & 1024 & 63.50 & 59.96 & \underline{73.18} & 88.87 & 82.35 & 11 & 4,096 / 5,741\\
& Pro & 1536 & 63.50 & 59.96 & \textbf{74.32} & 92.22 & 80.59 & 8 & 8,630 / 5,741\\
& OSWorld-G & 1024 & 68.82 & 76.47 & \underline{78.63} & 95.69 & 81.97 & 0 & 3,753 / 1,722\\
& OSWorld-G & 1536 & 68.82 & 76.47 & \textbf{79.80} & 98.43 & 81.08 & 1 & 5,699 / 1,722\\
\midrule
\multirow{8}{*}{4B}
& ScreenSpot & 1024 & 90.64 & 87.81 & \underline{88.84} & 96.46 & 91.28 & 7 & 3,840 / 2,734\\
& ScreenSpot & 1536 & 90.64 & 87.81 & \textbf{88.92} & 98.51 & 90.26 & 8 & 7,001 / 2,734\\
& V2 & 1024 & 93.16 & 90.64 & \underline{90.96} & 97.96 & 92.30 & 9 & 3,846 / 2,715\\
& V2 & 1536 & 93.16 & 90.64 & \textbf{91.51} & 99.14 & 92.31 & 6 & 6,987 / 2,715\\
& Pro & 1024 & 55.34 & 55.03 & \textbf{65.53} & 86.53 & 75.66 & 5 & 4,096 / 5,741\\
& Pro & 1536 & 55.34 & 55.03 & \underline{64.52} & 90.51 & 71.28 & 4 & 8,630 / 5,741\\
& OSWorld-G & 1024 & 57.45 & 56.67 & \underline{67.25} & 95.10 & 70.52 & 1 & 3,753 / 1,722\\
& OSWorld-G & 1536 & 57.45 & 56.67 & \textbf{67.65} & 98.43 & 68.73 & 0 & 5,699 / 1,722\\
\bottomrule
\end{tabular}
\end{table}

\section{Qualitative Comparisons and Spatial Readout}
\label{app:newcases}
We show eight successful PixelOne-Pro-35B-A3B predictions and visualize the model's token-to-pixel readout using saved outputs.

\subsection{Model-by-Model Comparisons}
\label{app:comparisoncases}
\begin{figure}[!htbp]
\centering
\captionsetup{width=\linewidth}
\includegraphics[width=\linewidth]{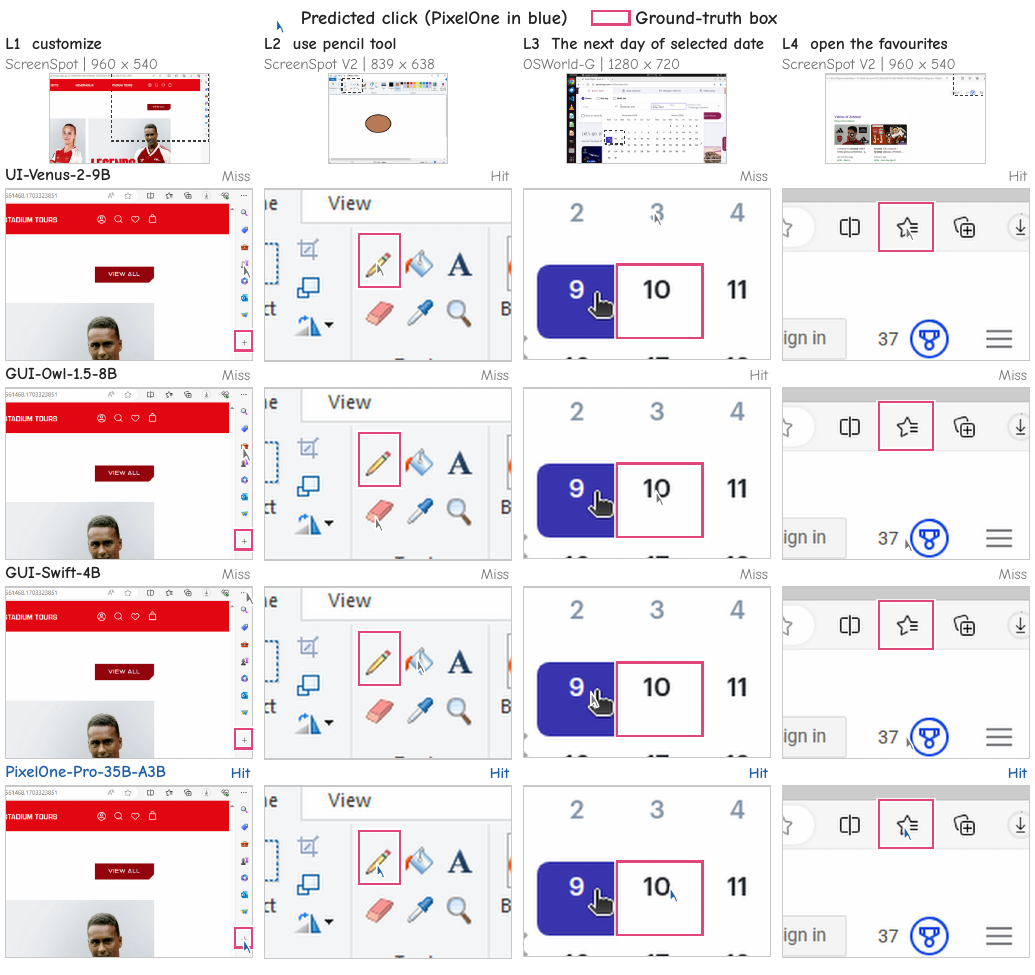}
\caption{\textbf{Successful grounding on lower-resolution screenshots.} These four original screenshots have a long side of at most 1,600 pixels. The top row shows the full screenshot with the display crop outlined. Subsequent rows show saved predictions on the same crop at the same scale within each column. Pink boxes denote targets, and mouse-pointer tips denote predictions. PixelOne-Pro pointers are blue, and baseline pointers are gray. Hit and Miss indicate the original point-in-box outcomes. All displayed predictions lie within their crops.}
\label{fig:comparisonlower}
\end{figure}

\begin{figure}[!htbp]
\centering
\captionsetup{width=\linewidth}
\includegraphics[width=\linewidth]{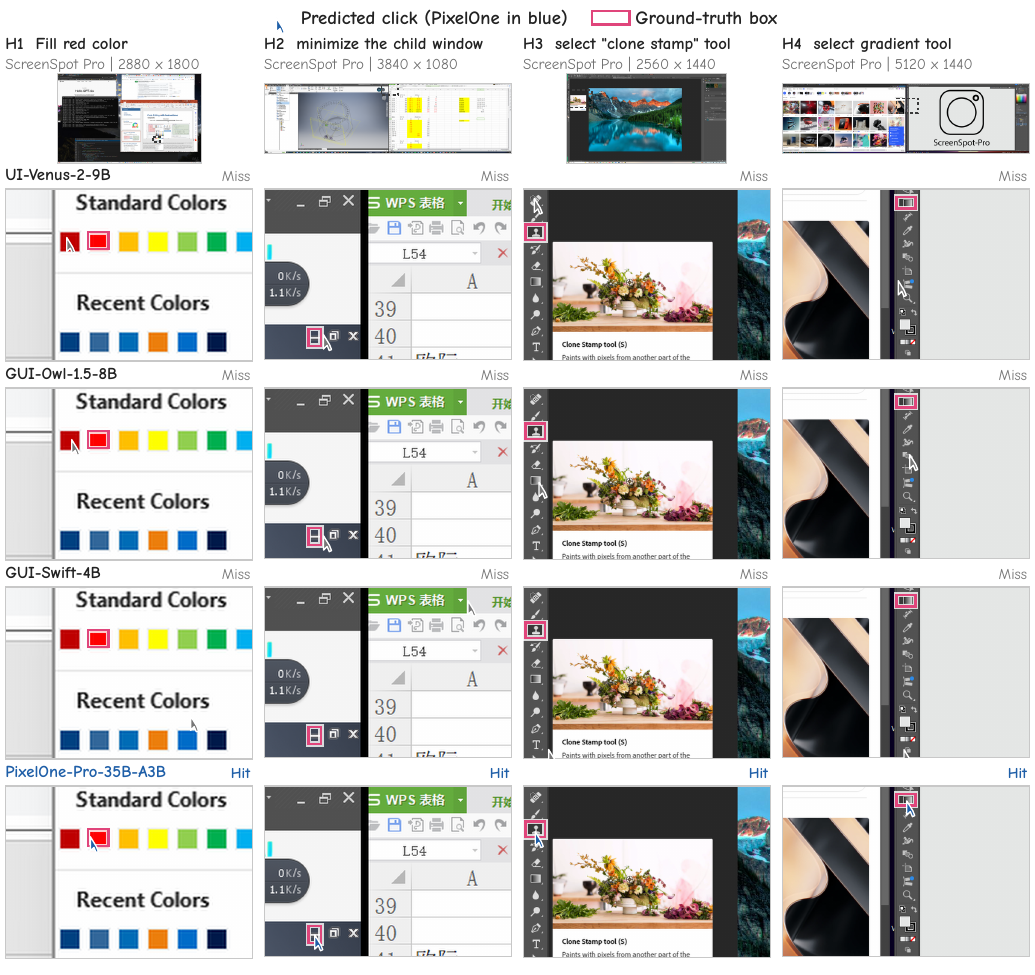}
\caption{\textbf{Successful grounding on higher-resolution screenshots.} Comparisons follow the column layout and visual conventions of \cref{fig:comparisonlower}. Original image sizes range from $2{,}560\times1{,}440$ to $5{,}120\times1{,}440$. Each model receives the same original screenshot and instruction through its evaluated input pipeline. Crops are for display only. These selected examples illustrate differences in click placement. Aggregate accuracy is reported in the main results.}
\label{fig:comparisonhigher}
\end{figure}

\Cref{fig:comparisonlower,fig:comparisonhigher} compare PixelOne-Pro with UI-Venus-2-9B, GUI-Owl-1.5-8B, and GUI-Swift-4B using the settings in \cref{app:recentbaselines}. Each column uses the same screenshot, instruction, target, and display crop for all models. PixelOne-Pro hits every selected target, while at least two displayed baselines miss each one.

\paragraph{Higher-resolution screenshots.}
\Cref{fig:comparisonhigher} shows screenshots with an original long side of at least 2,560 pixels, including wide multi-window interfaces. Targets include a small color swatch, a child-window control, and two editing tools. These are native benchmark images, not enlarged lower-resolution examples.

\subsection{From Token Scores to a Pixel Click}
\label{app:readoutcases}
\begin{figure}[!htbp]
\centering
\captionsetup{width=\linewidth}
\includegraphics[width=\linewidth]{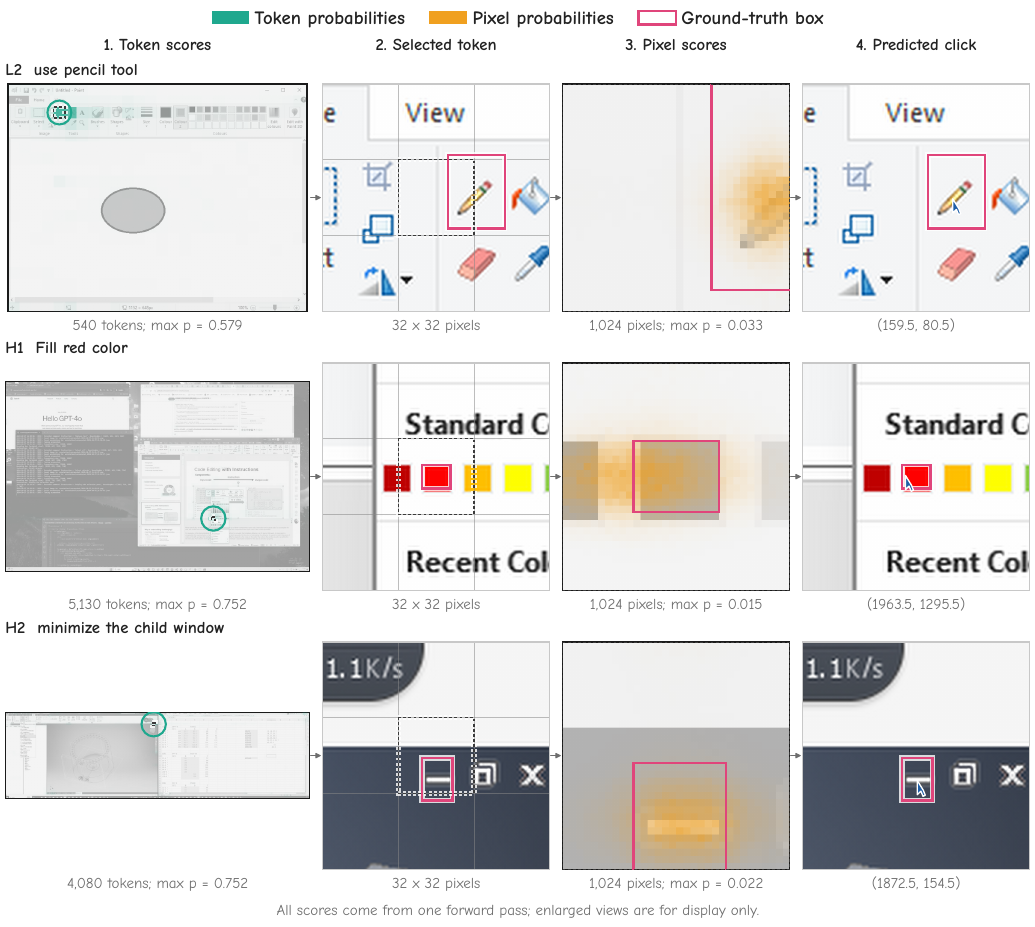}
\caption{\textbf{Token selection followed by within-token pixel selection.} The first column overlays full-image token probabilities and circles the selected-token location for visibility. The second enlarges that token and its immediate neighbors. Gray grid lines mark token boundaries, and a dashed outline marks the selected token. The third shows saved pixel probabilities within that $32\times32$ token. The last column marks the final click with a blue mouse pointer. Pink boxes are ground truth, not model inputs. Probability maps are normalized independently for display with their maxima labeled. Click coordinates are in original-image pixels.}
\label{fig:readoutcases}
\end{figure}

\Cref{fig:readoutcases} shows the saved scores for L2, H1, and H2. The decision head scores all visual tokens and selects the highest-scoring one. The pixel head scores valid pixels within that token, and its argmax gives the click. Teal and amber overlays show token and within-token pixel softmax distributions, respectively. Both stages come from one forward pass, not separate inference calls.

\end{document}